\documentclass[11pt]{article}

\usepackage[final]{acl}

\usepackage{times}
\usepackage{latexsym}

\usepackage[T1]{fontenc}

\usepackage[utf8]{inputenc}

\usepackage{microtype}

\usepackage{inconsolata}

\usepackage{graphicx}
\usepackage{booktabs}
\usepackage{natbib}
\usepackage{pifont}
\usepackage{amsmath}
\usepackage{amssymb}
\usepackage{multirow}
\usepackage{enumitem}
\usepackage{tikz}
\usepackage[dvipsnames]{xcolor}
\usepackage[table]{xcolor} 
\usepackage{subcaption}
\usepackage{pifont}
\usepackage{bm}
\usepackage{makecell}
\usepackage{tabularray}
\usepackage{hhline}

\definecolor{hlblue}{RGB}{210, 230, 250}  
\definecolor{hlgreen}{RGB}{230, 245, 230} 

\definecolor{my_green}{RGB}{0,155,85}   
\definecolor{my_blue}{RGB}{70,134,243}   
\definecolor{my_orange}{RGB}{230,136,98}   
\definecolor{my_purple}{RGB}{162,57,102}     
\definecolor{my_brown}{RGB}{130,25,0}     

\newcommand{\circled}[2]{%
  \begin{tikzpicture}[baseline=(char.base)]
    \node[shape=circle, fill=#1, inner sep=1pt, minimum size=1em] (char) 
    {\textcolor{white}{\sffamily\small{#2}}};
  \end{tikzpicture}%
}

\title{GeoLaux: A Benchmark for Evaluating MLLMs' \underline{Geo}metry Performance on \underline{L}ong-Step Problems Requiring \underline{Aux}iliary Lines}

\author{
  \textbf{Yumeng Fu\textsuperscript{1,2}},
  \textbf{Jiayin Zhu\textsuperscript{1,2}},
  \textbf{Lingling Zhang\textsuperscript{1,2}}\thanks{Corresponding authors.},
  \textbf{Wenjun Wu\textsuperscript{1,2}}\footnotemark[1],
  \textbf{Bo Zhao\textsuperscript{1,2}},
  \\
  \textbf{Shaoxuan Ma\textsuperscript{4}},
  \textbf{Yushun Zhang\textsuperscript{1,2}},
  \textbf{Jun Liu\textsuperscript{1,3}}
\\
  \textsuperscript{1}School of Computer Science and Technology, Xi'an Jiaotong University\\
  \textsuperscript{2}Ministry of Education Key Laboratory of Intelligent Networks and Network Security, China\\
  \textsuperscript{3} Shaanxi Province Key Laboratory of Big Data Knowledge Engineering, China\\
  \textsuperscript{4}School of Software Engineering, Xi'an Jiaotong University
\\
    \texttt{yumfuu@stu.xjtu.edu.cn, nickjunwork@163.com, \{zhanglling, liukeen\}@xjtu.edu.cn}
}

\begin{document}
\maketitle
\begin{abstract}
Geometry problem solving (GPS) poses significant challenges for Multimodal Large Language Models (MLLMs) in diagram comprehension, knowledge application, long-step reasoning, and auxiliary line construction. However, current benchmarks lack fine-grained evaluation for long-step problems necessitating auxiliary construction.
To address these limitations, we present GeoLaux, a fine-grained annotated dataset comprising 2186 calculation and proof problems. It features long-step reasoning (with an average solution length of 6.51 steps, maximum of 24 steps) and auxiliary line construction (required in 41.8\% of problems).
Building on the dataset, we conduct a comprehensive five-dimensional evaluation of 23 leading MLLMs. The evaluation yields three pivotal findings: 
First, models perform significantly worse on long-step problems compared to short-step ones, with 18 models exhibiting a performance drop of over 50\%. 
Second, it is crucial to enhance models' understanding, awareness, and proficiency in auxiliary line construction, which is vital for overall geometric reasoning. 
Third, limited answer hints effectively improve process correctness, whereas explicit answers lead models to neglect intermediate reasoning steps.
These findings position GeoLaux both to benchmark MLLMs geometry reasoning abilities and to guide their improvement. 
Data and code are available at \url{https://github.com/Candice-yu/GeoLaux}
\end{abstract}
\begin{figure}[t]
\centering
\includegraphics[width=0.95\columnwidth]{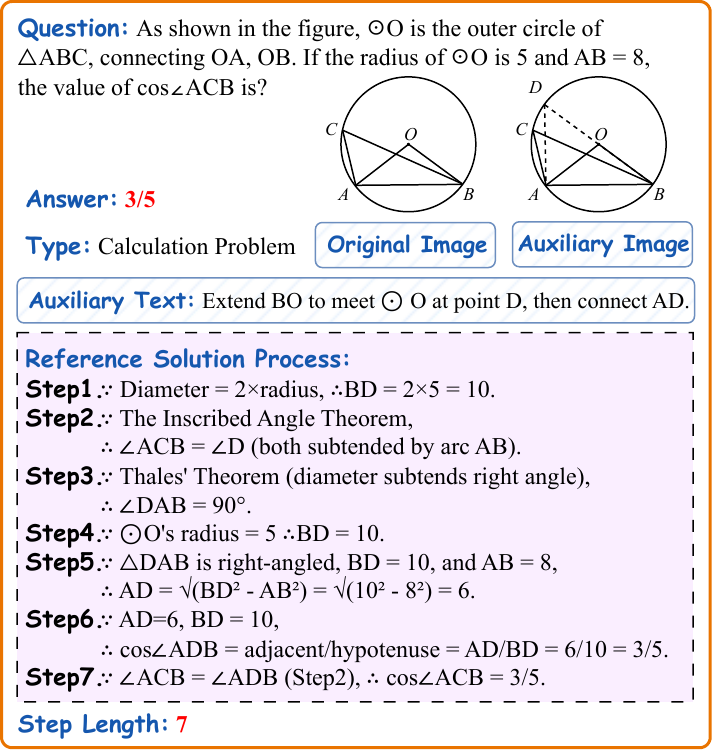} %
\caption{An illustration of example from GeoLaux.}
\label{fig:data_example}
\end{figure}

\section{Introduction}
\label{sec:Introduction}

As a cornerstone of mathematics, Geometry Problem Solving (GPS) epitomizes the advanced cognitive patterns of the human mind. It demands the integration of diverse core competencies: extensive geometric knowledge base, rigorous logical reasoning, precise computational skills, spatial visualization ability, and strategic auxiliary line construction \citep{jonsson2022creative,yan2025survey}. Given these challenging requirements, this task has consistently attracted widespread attention from the community \citep{lu2021inter,chen2021geoqa,wu2024gps,trinh2024solving,cheng2025geouni}.

\begin{table*}[t]
    \centering
    \setlength{\tabcolsep}{6.8pt}
    \setlength{\aboverulesep}{0.6pt}         
    \setlength{\belowrulesep}{0.6pt}         
    \scalebox{0.75}{
    \begin{tabular}{l|ccccccc|ccc}
    \toprule
    \multirow{2}{*}{Benchmark} & \multirow{2}{*}{Size} & \multirow{2}{*}{Type} & \multirow{2}{*}{\makecell[b]{Steps Length\\Avg. / Max}} & \multirow{2}{*}{\makecell[b]{Long-step \\ Prob. Num.}} & \multirow{2}{*}{\makecell[b]{Ultra-long \\ Prob. Num.}} & \multirow{2}{*}{\makecell[b]{Auxiliary \\ Lines}}  & \multirow{2}{*}{Source} & \multicolumn{3}{c}{Evaluation} \\
    \hhline{~|~~~~~~~:---}
    & & & & & & & & Answer & Error & Step \\
    \midrule
    Geometry3K \shortcite{lu2021inter} &601& \circled{my_purple}{C} & - & -& -& \textcolor{BrickRed}{$\times$} & \circled{my_green}{S} & \textcolor{ForestGreen}{$\boldsymbol{\checkmark}$} &\textcolor{BrickRed}{$\times$} &\textcolor{BrickRed}{$\times$}\\
    GeoQA \shortcite{chen2021geoqa}&755& \circled{my_purple}{C}   & 1.96 / 4 & 0& 0 & \textcolor{BrickRed}{$\times$} & \circled{my_green}{S} & \textcolor{ForestGreen}{$\boldsymbol{\checkmark}$} &\textcolor{BrickRed}{$\times$} &\textcolor{BrickRed}{$\times$}\\
    UniGeo \shortcite{chen2022unigeo}&1447 & \circled{my_purple}{C}\circled{my_purple}{P}  & - & -& - &\textcolor{BrickRed}{$\times$}  & \circled{my_green}{S} & \textcolor{ForestGreen}{$\boldsymbol{\checkmark}$} &\textcolor{BrickRed}{$\times$} &\textcolor{BrickRed}{$\times$}\\
    PGPS9K \shortcite{zhang2023multi} &1000 & \circled{my_purple}{C} & 2.43 / - & 0& 0&\textcolor{BrickRed}{$\times$}  & \circled{my_blue}{P}\circled{my_green}{S} & \textcolor{ForestGreen}{$\boldsymbol{\checkmark}$} &\textcolor{BrickRed}{$\times$} &\textcolor{BrickRed}{$\times$}\\
    IMO-AG-30 \shortcite{trinh2024solving} &30 & \circled{my_purple}{P}  & -& -& - & \textcolor{ForestGreen}{$\boldsymbol{\checkmark}$} & \circled{my_green}{S} & \textcolor{ForestGreen}{$\boldsymbol{\checkmark}$} &\textcolor{BrickRed}{$\times$} &\textcolor{BrickRed}{$\times$}\\
    GPSM4K \shortcite{anand2024improving} &200 & \circled{my_purple}{C}\circled{my_purple}{P} & -& -& - & \textcolor{BrickRed}{$\times$} & \circled{my_green}{S}\circled{my_orange}{A} & \textcolor{ForestGreen}{$\boldsymbol{\checkmark}$} &\textcolor{BrickRed}{$\times$} &\textcolor{BrickRed}{$\times$}\\
    GeoEval \shortcite{zhang2024geoeval} &2000 & \circled{my_purple}{C}  & -& -& - & \textcolor{BrickRed}{$\times$} & \circled{my_blue}{P}\circled{my_orange}{A}\circled{my_green}{S} & \textcolor{ForestGreen}{$\boldsymbol{\checkmark}$} &\textcolor{ForestGreen}{$\boldsymbol{\checkmark}$} &\textcolor{BrickRed}{$\times$}\\
    GeoSense \shortcite{xu2025geosense} &1789 & \circled{my_purple}{C}  & 5.70 / 16*& 169*& 14* & \textcolor{BrickRed}{$\times$} & \circled{my_blue}{P}\circled{my_green}{S} & \textcolor{ForestGreen}{$\boldsymbol{\checkmark}$} &\textcolor{ForestGreen}{$\boldsymbol{\checkmark}$} &\textcolor{BrickRed}{$\times$}\\
    SolidGeo \shortcite{wang2025solidgeo} &3113 & \circled{my_purple}{C}  & - & -& - & \textcolor{BrickRed}{$\times$}& \circled{my_green}{S}\circled{my_blue}{P} & \textcolor{ForestGreen}{$\boldsymbol{\checkmark}$} &\textcolor{ForestGreen}{$\boldsymbol{\checkmark}$} &\textcolor{BrickRed}{$\times$}\\
    GeoLaux (ours) &2186& \circled{my_purple}{C}\circled{my_purple}{P}  & 6.51 / 24& 292& 208 & \textcolor{ForestGreen}{$\boldsymbol{\checkmark}$} & \circled{my_green}{S} & \textcolor{ForestGreen}{$\boldsymbol{\checkmark}$} &\textcolor{ForestGreen}{$\boldsymbol{\checkmark}$} &\textcolor{ForestGreen}{$\boldsymbol{\checkmark}$}\\
    \bottomrule
    \end{tabular}
    }
    \caption{Comparison with other geometry benchmarks. \textbf{Type:} \protect\circled{my_purple}{C}=\underline{C}alculation, \protect\circled{my_purple}{P}=\underline{P}roving. \textbf{Source:} \protect\circled{my_green}{S}=\underline{S}elf-Sourced, \protect\circled{my_blue}{P}=Collected from \underline{P}ublic Datasets, \protect\circled{my_orange}{A}=\underline{A}ugmented from Existing Data. *: Derived from the official dataset as originally textually unreported. Long-step problems: 9–12 steps. Ultra-long problems: over 13 steps.}
    \label{tab:benchemark}
\end{table*} 

Multimodal Large Language Models (MLLMs), represented by GPT-4o \cite{hurst2024gpt}, have recently emerged as a significant focus of research attention. By combining the inherent strengths of Large Language Models (LLMs) in knowledge, reasoning, and calculation with visual modules, MLLMs demonstrate remarkable performance in multimodal reasoning  \citep{Driess2023PaLMEAE, yin2024survey}. These characteristics reflect their potential for GPS, leading to numerous studies \citep{anand2024geovqa, zhang2024geoeval} evaluating MLLMs' geometric reasoning abilities.

Table \ref{tab:benchemark} summarizes existing benchmarks for evaluating MLLMs' geometric reasoning, which generally exhibit three main limitations:
\textit{(1) Absence of long-step reasoning evaluation.} Long-step reasoning is essential for solving complex geometric problems, making accurate long-step evaluation crucial for advancing model performance. However, constrained by limited step length, current benchmarks cannot fully assess MLLMs' long-step reasoning, illustrated by SolidGeo \citep{wang2025solidgeo} having only 6.7\% multi-step questions and GeoSense \citep{xu2025geosense} with just 14 problems exceeding 13 steps.
\textit{(2) Absence of auxiliary line evaluation.} 
Constructing correct auxiliary lines is a vital evaluation dimension that critically tests MLLMs' deep understanding of both geometric diagrams and textual problems. Given an image with $n$ geometric primitives (i.e. points, lines and circles), there exist $n^3$ possible auxiliary line constructions \citep{marinkovic2017argotrics}, posing a significant challenge to spatial reasoning capabilities of models. However, benchmarks evaluating this aspect are currently lacking.  
\textit{(3) Coarse-grained process evaluation.}  While recent LLM research highlights the necessity of fine-grained trajectory analysis (e.g., error localization and quality assessment) for evaluating long-step reasoning \cite{zhang2025openprm, yan2026tide, xu2026odysseyarena}, existing GPS benchmarks solely use answer correctness as the success criterion, with process analysis limited to error classification \citep{zhang2024geoeval, xu2025geosense}. Such coarse-grained evaluation fails to identify model weaknesses and guide improvement.

To address these, we present a plane geometry problem dataset GeoLaux, which comprises 2186 problems collected from Zhongkao mathematics papers across 34 provincial-level regions in China over past two years. This dataset exhibits three key characteristics: (1) \textbf{long-step reasoning} with problems averaging 6.51 solution steps (up to 24 steps), (2) \textbf{annotated auxiliary lines} including both detailed construction methods and resulting geometric diagrams, and (3) \textbf{dual problem types} comprising 1,418 calculation and 768 proving problems. As shown in Figure \ref{fig:data_example}, we annotate step-by-step solution process for each problem, establishing foundation for fine-grained process evaluation.

Besides the dataset, we designe a fine-grained framework to evaluate MLLM problem-solving. Specifically, beyond general \textbf{answer correctness} (measured by ACS) and \textbf{error type}, we introduce three additional dimensions tailored for complex reasoning: \textbf{solution process correctness} (PCS), \textbf{solution process quality} (PQS), and \textbf{auxiliary line construction}, totaling 5 dimensions and 3 metrics. Leveraging this evaluation framework, we assess 23 state-of-the-art MLLMs, including 10 thinking models and 13 non-thinking models. The results demonstrate that Gemini-2.5 Pro \citep{gemini2025Pro} achieves the highest overall performance, followed by o3 \citep{o32025} and Qwen3-VL-32B-thinking \citep{yang2025qwen3}. Our analysis reveals three critical findings:
\begin{itemize}[nosep,leftmargin=*, labelsep=0.2em, itemsep=0pt]
    \item\textbf{MLLM performance degrades significantly on long-step problems compared to shorter ones:} Nine models show a performance drop of over 50\%  from short-step to ultra-long-step problems, with some exceeding 90\% or even reaching 100\%.
    \item \textbf{MLLMs lack the awareness and capability to proactively construct auxiliary lines:} They fail to recognize when such constructions are necessary and struggle to generate correct lines when attempted. Yet, their strong performance gains under standard construction prompting underscore the vital role of accurate auxiliary lines.
    \item \textbf{MLLMs perform better with limited answer hints, but decline when answers are clearly disclosed:} 
    Compared to free-response problems, models often achieve higher process correctness scores on multiple-choice questions, yet lower on proving problems. This implies options act as incentives for reasoning, while explicit answers lead models to disregard process correctness.
\end{itemize}

In this way, we establish a fair evaluation benchmark that not only assesses MLLMs' reasoning capabilities on long-step auxiliary line problems, but also provides clear guidance for enhancing their geometry reasoning performance.

\section{Related Work}
\label{relatedwork}
Prior to the rapid development of MLLMs, several established benchmarks existed for evaluating traditional geometric problem solving methods, including Geometry3K \citep{lu2021inter}, GeoQA \citep{chen2021geoqa}, and UniGeo \citep{chen2022unigeo}. Typically featuring low difficulty and a limited variety of problem types, they are inadequate for meeting current evaluation demands.
Consequently, several specialized benchmarks for evaluating MLLMs' math reasoning capabilities have emerged in recent years, such as MathVista \citep{lu2023mathvista}, MathVerse \citep{zhang2024mathverse}, We-Math \citep{qiao2025we}, but these works have not focused on analyzing the aspect of GPS. Among works focused on geometry, IMO-AG-30 \citep{trinh2024solving} provide 30 Olympiad-level problems, but exclusively address theorem proving without algebra; GeoEval \citep{zhang2024geoeval} restructures existing problems into a unified format to assess answer accuracy; GeoSense \citep{xu2025geosense} evaluates recognition of geometric principles but overlooks key dimensions such as diagram comprehension and auxiliary line construction; SolidGeo \citep{wang2025solidgeo} focuses on solid geometry problems, yet lacks attention to critical auxiliary line construction and includes only 6.7\% multi-step questions. Similarly, while the recent Geoint benchmark \citep{wei2025geoint} introduces Lean 4 to formalize auxiliary lines, its evaluation relies on rigid code-level matching, falling short of deeply assessing the models' dynamic geometric reasoning capabilities.

In conclusion, current MLLM geometry benchmarks lack fine-grained process evaluation, auxiliary line assessment, and multi-step reasoning evaluation, necessitating new evaluation standards.

\section{GeoLaux DataSet}
\label{sec:dataset}
GeoLaux is a challenging plane geometry dataset comprising 2186 fully verified problems, divided into 1418 calculation and 768 proving problems. Of the calculation problems, 522 are multiple-choice (single-answer) and 896 are free-response questions.
As shown in Figure \ref{fig:data_example}, each problem in our dataset contains 8 annotated elements: problem text, geometric diagram, type (proving or calculation), answer, step-by-step solution, step length, auxiliary line construction text, and auxiliary line construction image.
This section describes its semi-automated construction pipeline, which consists of three main stages: data acquisition, step segmentation, and auxiliary line extraction.

\paragraph{Data Acquisition.}
To ensure data authenticity and comprehensiveness, we systematically select plane geometry problems from the High School Entrance Examination (HSEE/Zhongkao) mathematics papers across China's 34 provincial-level regions as our raw dataset. These questions comprehensively cover the core plane geometry knowledge required in secondary education.
Beyond the original problems' texts and diagrams, we further incorporate expert-curated standard answers and solutions from official exam materials. Every problem is carefully verified for: (1) diagram clarity, (2) text-diagram correspondence, (3) answer accuracy, and (4) detailed annotation of solution processes.
These validated real-world problems lay a solid foundation for our benchmark.

\begin{figure}[t]
\centering
\includegraphics[width=0.9\columnwidth]{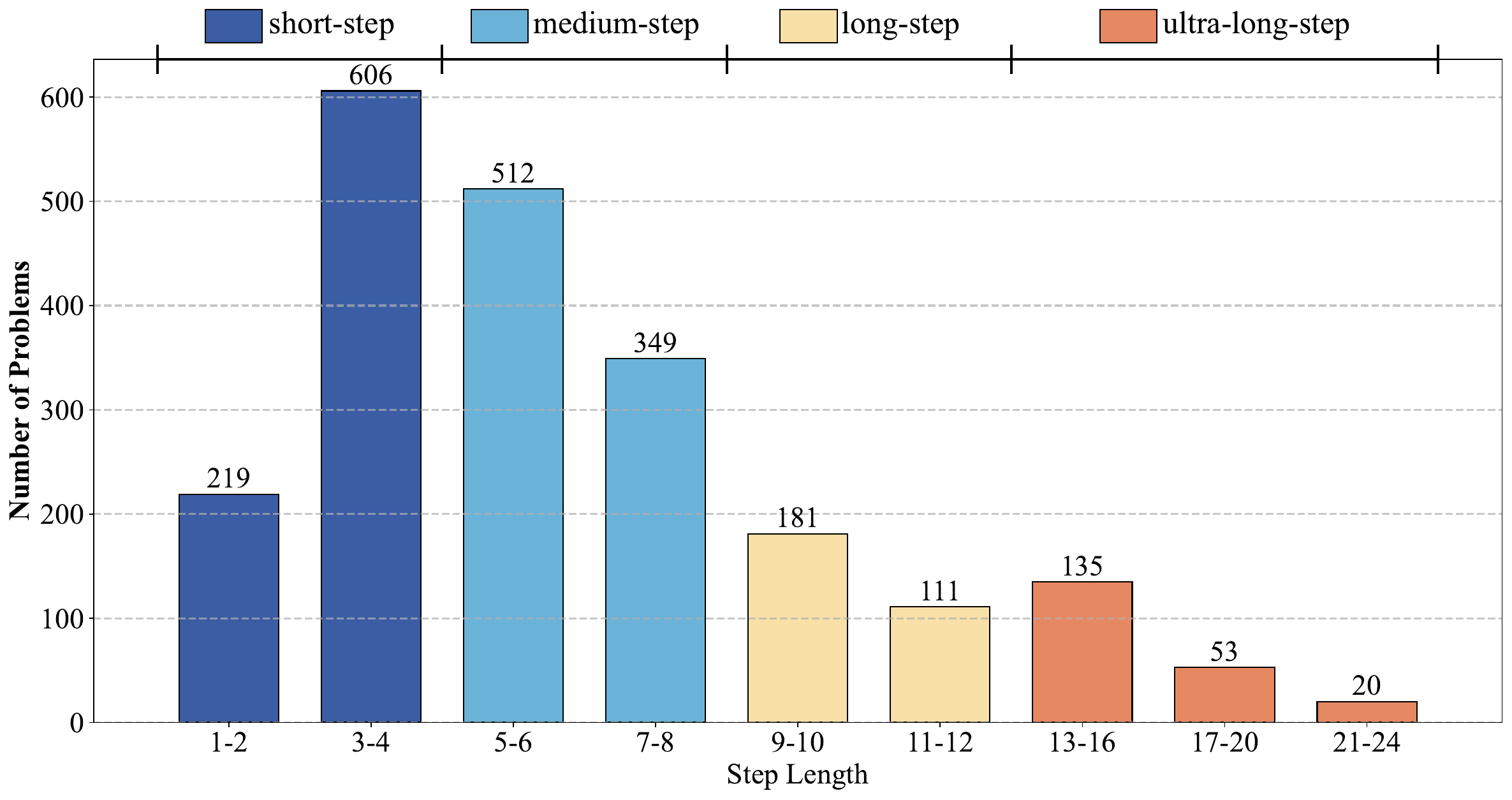} %
\caption{Problem quantity statistics across step lengths}
\label{fig:Step_Number}
\end{figure}

\paragraph{Step Segmentation.}
Based on the fundamental logical structure of mathematical reasoning, we define each "because($\because$)-therefore($\therefore$)" pair as a complete reasoning step. Gemini-2.5-Pro is tasked with segmenting the pre-annotated standard solutions into such steps while explaining each segmentation decision to make sure the splitting follows our rules. This process generates standardized solution step length for every problem in our dataset, serving as crucial labels for subsequent analysis. To ensure the authenticity and accuracy of these step counts, we conducted a rigorous manual review of the model-segmented results. 
Figure \ref{fig:Step_Number} presents the step length distribution of all 2186 problems. The dataset includes a substantial number of long-step and ultra-long-step problems, with an average step of 6.51 and a maximum step of 24. This step segmentation process establishes an ideal testbed for evaluating long-step reasoning capabilities.

\begin{figure}[t]
\centering
\includegraphics[width=0.77\columnwidth]{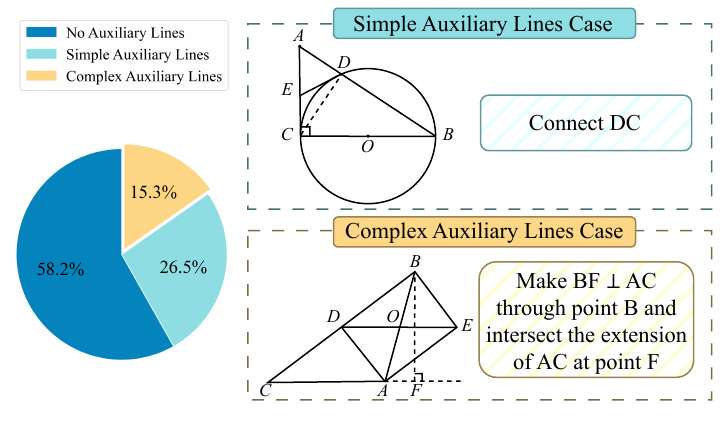} %
\caption{Distribution of auxiliary line types.}
\label{fig:Auxiliary_distribution}
\end{figure}

\paragraph{Auxiliary Line Extracting.}
Our dataset contains numerous problems requiring auxiliary lines to evaluate MLLMs' capability in auxiliary structure construction. We employed Gemini-2.5-Pro to extract auxiliary line construction steps from the pre-annotated standard solutions. Considering the importance of visual input for MLLMs, we further manually collected corresponding diagrams with auxiliary lines from original exam papers, forming the visual-text pairs illustrated in Figure \ref{fig:Auxiliary_distribution}.
The auxiliary lines are classified by difficulty into simple ones (involving only point connections) and complex ones (creating new geometric primitives like perpendiculars, angle bisectors, or inscribed circles).
According to our statistics, GeoLaux includes 334 problems requiring complex auxiliary lines (15.3\% of the total) and 580 problems needing simple auxiliary lines (26.5\% of the total), laying the groundwork for comprehensive evaluation.

\paragraph{Comparison with Existing DataSets.}
As illustrated in Table \ref{tab:benchemark}, GeoLaux demonstrates three key advantages over other datasets: \textbf{(1) Long Steps:} GeoLaux surpasses existing benchmarks in both average and maximum solution step lengths. Notably, with 208 ultra-long-step problems (exceeding 13 steps), it poses significant challenges for models. \textbf{(2) Unique Auxiliary Line Annotation:} While auxiliary lines are essential for IMO-AG-30 \citep{trinh2024solving}, which contains only 30 problems and provides no annotations for auxiliary lines. GeoLaux is the first large benchmark to provide complete, explicit and multi-modal annotations for auxiliary line construction methods. \textbf{(3) Integrated Calculation and Proving Problems:} GeoLaux enables fair cross-type performance comparisons of MLLMs in geometric problem solving.

\section{Evaluation Strategy}
\label{sec:Evaluation Methods}
\begin{figure*}[t!]
\centering
\includegraphics[width=0.9\textwidth]{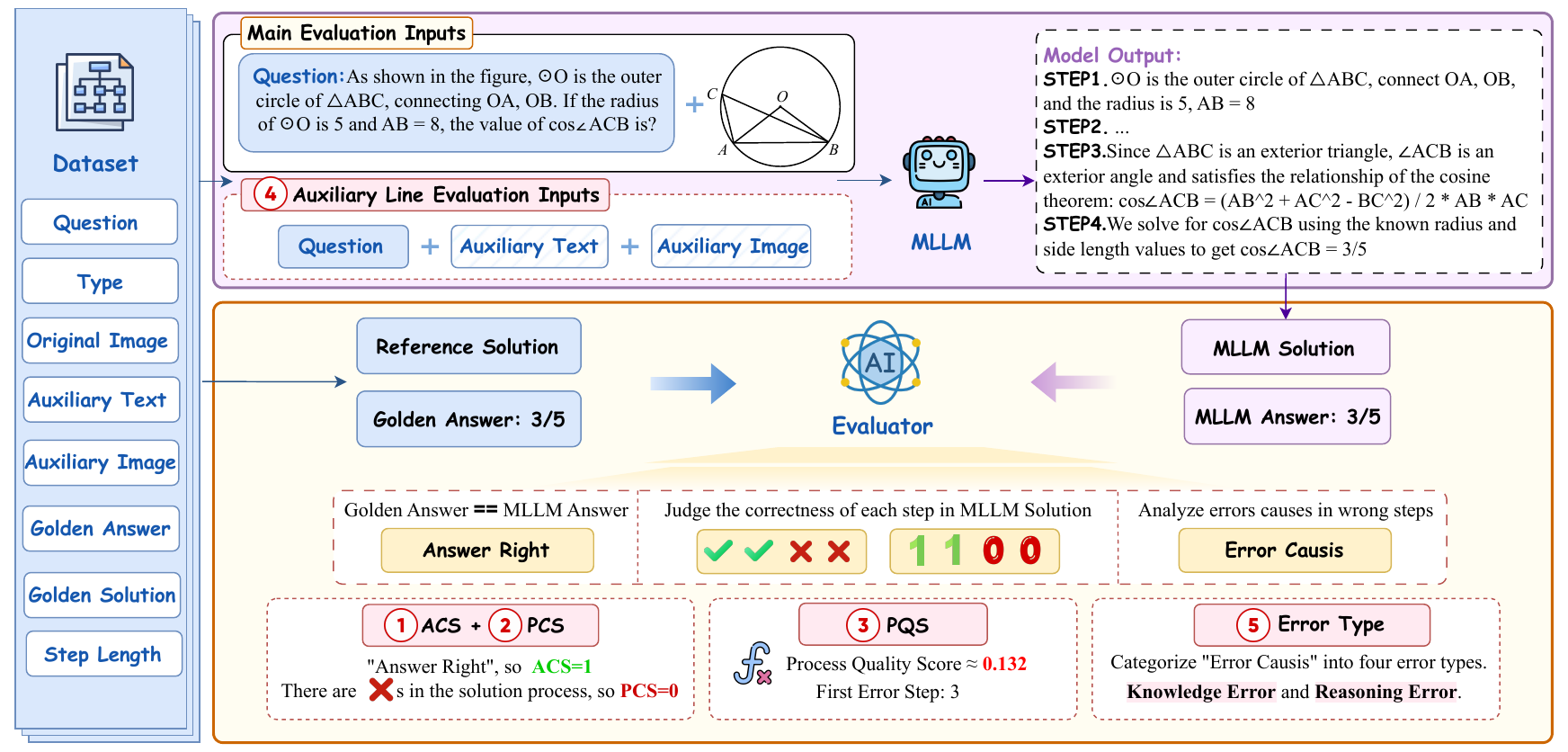} 
\caption{Five-dimension evaluation framework of GeoLaux. Given golden answer and solution from dataset, evaluator conducts a comprehensive assessment of MLLM outputs across the following dimensions: \ding{172} answer correctness, \ding{173} process correctness, \ding{174} process quality, \ding{175} auxiliary line impact, and \ding{176} error type.}
\label{fig:framework}
\end{figure*}

Based on the dataset, we propose a 5-dimensional evaluation framework (Figure \ref{fig:framework}). In addition to general \textbf{answer correctness} (with metric ACS) and \textbf{error type}, we devise 3 novel evaluation dimensions: \textbf{process correctness} (with metric PCS), \textbf{process quality} (with metric PQS), and \textbf{auxiliary line impact}. The following four subsections detail these evaluation dimensions, with a concluding analysis of the evaluation framework's reliability.

\subsection{Correctness Evaluation (ACS \& PCS)}
\label{correctness metric}
The first two dimensions both evaluate MLLMs' solution correctness, employing the Answer Correctness Score (ACS) alongside our novel and stricter Process Correctness Score (PCS). While ACS checks only the final answer, PCS mandates correct reasoning. To facilitate the independent evaluation of these two components, we require MLLMs to output step-by-step reasoning and the summarized final answer in JSON format, with the format validity ensured through manual review.

\paragraph{Answer Correctness Evaluation (ACS).} 
The model's self-summarized answer is then compared with the ground-truth answer through our evaluator model. Specifically, for problem $q$ with ground-truth answer $a$ and model answer $\hat{a}$, the Answer Correctness Score (ACS) is formally defined as:
\begin{equation}
\label{eq:acs}
\mathrm{ACS}= 
\begin{cases} 
1 & \text{if } \hat{a} = a \\ 
0 & \text{otherwise} 
\end{cases}
.
\end{equation}
Depending on the problem type, $a$ can be a numerical value, an option, or a geometric condition.

\paragraph{Process Correctness Evaluation (PCS).}
We observe that MLLMs occasionally generate correct answers through flawed processes, a phenomenon we term \textbf{False Positives}. This necessitates a stricter process evaluation to assess true performance. 

The structured step-by-step solutions enable our evaluator to score each individual reasoning step, assigning 1 for correct steps and 0 for incorrect ones. Given an n-step solution process, the evaluator assigns scores as follows:
\begin{equation}
\eta = (\eta_1, \eta_2, \ldots, \eta_n), \quad \eta_i \in \{0, 1\}.
\label{eq:01score}
\end{equation}
Building on this fine-grained scoring system, the Process Correctness Score (PCS) is defined as:
\begin{equation}
\label{eq:pcs}
\mathrm{PCS} = 
\begin{cases} 
1 & \text{if } (\hat{a} = a )\ \land \ (0 \notin \eta) \\ 
0 & \text{otherwise} 
\end{cases}
.
\end{equation}
This metric rigorously evaluates problem-solving correctness, requiring not only accurate final answers but also error-free reasoning processes.

\subsection{Process Quality Evaluation (PQS)}
\label{sec:Fine-grained Process Evaluation}

To ensure a fair comparison of solution quality across different MLLMs, we design a step weight function that assigns specific weights to each step's score, ultimately computing a weighted overall process quality score. Our weighting function incorporates the following considerations:
\begin{enumerate}[nosep,leftmargin=*, labelsep=0.2em, itemsep=0pt]
    \item\textbf{Decreasing function:} Models that make errors in earlier steps exhibit weaker capability for accurate long‑step reasoning, which implies that earlier steps should be assigned higher weights.
    \item \textbf{Convex function:} The importance gap is larger for earlier steps and smaller for later ones. For example, two models erring at steps 2 and 4 should show larger score differences than those erring at steps 12 and 14.
    \item \textbf{Moderate decreasing rate:} Weighting function should not decrease too rapidly. For long-step problems, performance in later steps remains critical and should retain significant weight.
\end{enumerate}

Given these considerations, for a solution process with $n$ steps, we define the weight function for the $i$-th step as:
\begin{equation}
y_i = e^{-\frac{i}{n}}.
\label{eq:weight_function}
\end{equation}

The initial process quality score, using Equation \ref{eq:01score} for grading and Equation \ref{eq:weight_function} for weighting, is defined as follows:
\begin{equation}
\mathrm{PQS'} = \sum_{i=1}^n \frac{\eta_i \cdot y_i}{\sum_{j=0}^n y_j} 
  = \sum_{i=1}^n \frac{\eta_i \cdot e^{-\frac{i}{n}} }{\sum_{j=1}^n e^{-\frac{j}{n}}}.
\label{equation_initial_process_score}
\end{equation}

However, since model solutions always contain some correct steps, $\mathrm{PQS'}$ consistently falls between 0.6 and 1, failing to highlight models' differences in reasoning ability. Therefore, we apply the $\tanh$ activation function on $\mathrm{PQS'}$.
\begin{equation}
\mathrm{PQS} = \tanh\bigl(\alpha (\mathrm{PQS'} - 1)\bigr) + 1,
\label{equation_final_process_score}
\end{equation}

where $\alpha$ is a hyperparameter set to 3.5. Through this approach, we obtain the final $\mathrm{PQS}$ normalized to [0,1], 
which provides a more refined metric for evaluating MLLMs' reasoning capabilities. Appendix \ref{appendix_pqs} records our other considerations on the weight functions and hyperparameters.
\subsection{Auxiliary Line Evaluation}
\label{sec:Meta Auxiliary}
\begin{figure}[t]
\centering
\includegraphics[width=1\columnwidth]{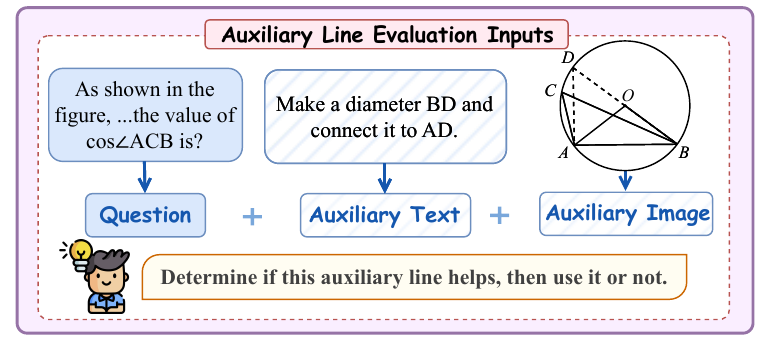} 
\caption{Illustration of auxiliary line evaluation inputs.}
\label{fig:Auxiliary_line_input}
\end{figure}

To rigorously evaluate the model's awareness of constructing auxiliary lines and verify their importance, we establish two distinct settings: one allows model to solve problems autonomously without additional prompts; the other, we provide MLLM with textual auxiliary line construction methods and the corresponding annotated diagrams from GeoLaux (as shown in Figure \ref{fig:Auxiliary_line_input}), heuristically prompting model to decide whether to adopt these lines to solve problems. Based on these two scenarios, our evaluation focuses on the following two aspects.

First, we examine model's auxiliary construction awareness through calculating construction rate under both scenarios. Specifically, construction rate is obtained by having our evaluator determine the probability of MLLM employing specific auxiliary line constructions through keyword detection in its solutions. By comparing construction rates between autonomous and prompted scenarios, we can quantify model's inherent awareness of utilizing auxiliary lines.

Second, we measure the impact of auxiliary lines through calculating model's PCS after auxiliary construction under both scenarios. By comparing the PCS scores across these two settings, we can analyze how proper auxiliary construction influences the geometric problem-solving performance.
\subsection{Error Type Evaluation.}
\label{sec:error_eva}
Understanding error causes enables targeted improvements for MLLMs in GPS. Consequently, we additionally conduct detailed error analysis for each step in the model's solution process, categorizing failure steps into four distinct types: \textbf{figure comprehension error, knowledge error, calculation error, and logical reasoning error}.

Figure comprehension error means model fails to correctly comprehend geometric elements and their relationships in the geometric figure. Knowledge error arises when the model applies incorrect formulas, theorems, or properties. Calculation error refers to mistakes in numerical calculations. And logical reasoning error encompasses flaws in the deductive process, including invalid causal relationships, over-skipping of reasoning steps, taking groundless assumptions as fact, etc. 

\subsection{Evaluation Framework Reliability}
In our evaluation framework, after MLLM generates responses based on question and diagram, both its reasoning process and reference process from dataset are fed into the evaluator. 
Evaluator performs a relatively simple task where it scores each solution step by checking whether it contains errors (as defined in section \ref{sec:error_eva}), assigning 0 if an error is present and 1 otherwise.
Notably, our evaluation framework does not employ a step-matching logic. The reference solution serves as an evaluation aid rather than the only correct solution path. Therefore, valid alternative approaches, such as adopting a coordinate geometry approach instead of the reference's auxiliary line method or utilizing different yet effective auxiliary constructions, are not penalized. The application of key knowledge, computations, and handling of geometric primitives in reference solutions helps reduce error misjudgments, thereby enhancing evaluator's reliability.

\begin{table}[t]
\setlength{\tabcolsep}{6pt}
    \scalebox{0.7}{
    \begin{tabular}{l|cc}
        \toprule
        \textbf{Evaluator Model} & \textbf{Answer Acc.} & \textbf{Step Acc.} \\
        \midrule
        o4-mini & 99.6\% & 97.2\% \\
        o4-mini (w/o Reference) & 98.5\% & 94.9\% \\
        Gemini-2.5-Pro & 99.2\% & 94.1\% \\
        Gemini-2.5-Pro (w/o Reference) & 98.7\% & 91.7\% \\
        Claude-4.5 & 95.8\% & 89.7\% \\
        GPT-4o & 93.8\% & 82.3\% \\
        \bottomrule
    \end{tabular}
    }
\caption{Performance of different evaluator models.}
\label{tab:evaluator}
\end{table}
\begin{table}[t]
\scalebox{0.62}{
\begin{tabular}{ll}

\toprule

\textbf{Institutions} & \textbf{MLLMs} \\
\midrule
\multirow{2}{*}{Deepmind} & Gemini-2.0-Flash-Thinking$^\dagger$ \shortcite{gemini2024thinking}, \\
                                 & Gemini-2.5-Pro$^\dagger$ \shortcite{gemini2025Pro}  \\
\cmidrule{2-2} 

\multirow{3}{*}{OpenAI} & GPT-4o \shortcite{hurst2024gpt}, GPT-4.1 \shortcite{gpt412025},  \\
                        & o1$^\dagger$ \shortcite{o12025}, o3-mini$^\dagger$ \shortcite{o3mini2025},\\
                        & o3$^\dagger$ \shortcite{o32025}, o4-mini$^\dagger$ \shortcite{o4mini2025}  \\
                        
\cmidrule{2-2}
\multirow{2}{*}{Anthropic}  & Claude3.7 \shortcite{claude372025}, Claude-4\shortcite{claude42025},\\ 
                            &Claude-4.5,Claude-4.5-Thinking$^\dagger$\shortcite{claude452025}\\

\cmidrule{2-2}
\multirow{3}{*}{Alibaba}            & QvQ-72B$^\dagger$ \shortcite{team2024qvq}, \\
                             & Qwen2.5-VL-7B,
                              Qwen2.5-VL-72B \shortcite{bai2025qwen2}  \\
                              &Qwen3-VL-32B, Qwen3-VL-32B-Thinking$^\dagger$ \shortcite{yang2025qwen3}  \\
\cmidrule{2-2}
Shanghai                        & InternVL3-8B, InternVL3-78B \shortcite{zhu2025internvl3}\\
AI Lab                        & InternVL3.5-8B, InternVL3.5-38B \shortcite{wang2025internvl3}  \\
\cmidrule{2-2}
Zhipu                       &GLM-4.1V-9B-Base, GLM-4.1V-9B-Thinking$^\dagger$ \shortcite{hong2025glm}\\
\bottomrule
\end{tabular}
}
\caption{Evaluated models and corresponding institutions. MLLMs marked with $^\dagger$ are thinking models.}
\label{tab:models_and_institutions}
\end{table}
\begin{table*}[h!]
    \centering
    \setlength{\tabcolsep}{0.4mm}
\scalebox{0.78}{   
\begin{tabular}{@{} l wc{1.8cm} | *{2}{wc{1.1cm}} | *{2}{wc{1.1cm}} | *{2}{wc{1.1cm}} | *{2}{wc{1.1cm}} | wc{1.8cm} | *{3}{wc{1.1cm}}  @{}}
    \toprule
    \multirow{2}{*}{\textbf{Model}}&\multirow{2}{*}{\textbf{Dataset}}&  \multicolumn{2}{c|}{$1-4$ Steps} & \multicolumn{2}{c|}{$5-8$ Steps} & \multicolumn{2}{c|}{$9-12$ Steps} & \multicolumn{2}{c|}{$13-24$ Steps} & \multicolumn{1}{c|}{Step-wise} & \multicolumn{3}{c}{\textbf{Overall AVG}} \\ 
    \cmidrule(lr){3-4} \cmidrule(lr){5-6} \cmidrule(lr){7-8} \cmidrule(lr){9-10} \cmidrule(lr){11-11} \cmidrule(lr){12-14} 
    & & ACS & PCS & ACS & PCS & ACS & PCS & ACS & PCS & $\Delta$PCS(\%) & ACS & PCS & PQS \\ 
    \midrule
    \multicolumn{14}{c}{\textbf{Thinking MLLMs}} \\ 
    \midrule
    QvQ-72B&all & 69.6 & 22.4 & 52.7 & \phantom{0}6.6 & 27.4 & \phantom{0}1.7 & 14.0 & \phantom{0}1.2 &  94.6 & 52.1 & 11.4 & 21.0 \\
    o3-mini &all& 60.7 & 21.8 & 54.7 & 13.0 & 39.7 & 12.7 & 16.5 & \phantom{0}5.5 &  74.8 & 51.8 & 15.7 & 27.2 \\
    GLM-4.1V-9B-T* &all& 91.1 & 71.0 & 78.6 & 44.3 & 68.3 & 31.7 & 35.7 & 10.7 &  84.9 & 75.8 & 42.8 & 57.5 \\
    o1 &mini & 86.3 & 64.5 & 80.9 & 57.3 & 82.9 & 61.0 & 42.9 & 35.7 &  44.7 & 79.7 & 58.8 & 68.6 \\
    Gemini-2.0-T* &all& 89.7 & 72.2 & 81.6 & 53.4 & 64.7 & 34.9 & 40.2 & 17.7 &  75.5 & 78.7 & 54.9 & 72.9 \\
    Claude-4.5-T* &all& 89.3 & 69.8 & 84.6 & 52.5 & 71.2 & 46.2 & 51.2 & 33.5 &  52.0 & 81.7 & 56.5 & 73.2 \\
    o4-mini&all& 94.5 & 78.3 & \multicolumn{1}{>{\columncolor{hlblue}}c}{94.0} & 70.3 & 91.4 & 70.2 & \multicolumn{1}{>{\columncolor{hlblue}}c}{81.1} & \multicolumn{1}{>{\columncolor{hlblue}}c}{63.4}& \multicolumn{1}{>{\columncolor{hlblue}}c}{19.0} & \multicolumn{1}{>{\columncolor{hlblue}}c}{92.8}& 72.9 & 81.1 \\
    Qwen3-VL-32B-T*&all & 94.9 & \cellcolor{hlblue}86.2 & 92.8 & 74.4 & 84.9 & 66.4 & 64.0 & 48.8 &  43.4 & 90.3 & 75.7 & 84.2 \\
    o3 &mini& 94.4 & 83.9 & 93.9 & \multicolumn{1}{>{\columncolor{hlblue}}c}{80.9} & \multicolumn{1}{>{\columncolor{hlblue}}c}{92.7} & 73.2 & 78.6 & 53.6 &  36.1 & 92.4 & \multicolumn{1}{>{\columncolor{hlblue}}c}{78.5} & {86.0}\\
    Gemini-2.5-Pro &all& \multicolumn{1}{>{\columncolor{hlblue}}c}{95.3} & 85.9 & 92.2 & 76.2 & 88.0 & \multicolumn{1}{>{\columncolor{hlblue}}c}{76.0} & 71.3 & 50.0 &  41.8 & 91.2 & 77.8 & \cellcolor{hlblue}88.6 \\
    \midrule
    \multicolumn{14}{c}{\textbf{Non-Thinking MLLMs}} \\
    \midrule
    Qwen2.5-VL-7B &all& 38.7 & \phantom{0}8.5 & 23.6 & \phantom{0}1.2 & 12.3 & \phantom{0}0.0 & \phantom{0}8.5 & \phantom{0}0.0 &  100.0\phantom{0} & 26.4 & \phantom{0}3.7 & 11.1 \\
    GPT-4o &all& 57.7 & 14.7 & 49.4 & \phantom{0}2.8 & 28.1 & \phantom{0}0.3 & 10.4 & \phantom{0}0.6 &  95.9 & 46.1 & \phantom{0}6.7 & 20.4 \\
    InternVL3-8B &all& 72.8 & 30.2 & 50.2 & \phantom{0}7.7 & 25.0 & \phantom{0}0.6 & 10.4 & \phantom{0}0.3 &  99.0 & 51.6 & 14.5 & 24.9 \\
    GLM-4.1V-9B &all& 69.2 & 31.8 & 43.4 & \phantom{0}8.1 & 20.2 & \phantom{0}1.0 & \phantom{0}9.1 & \phantom{0}0.0 &  100.0\phantom{0} & 46.8 & 15.4 & 26.1 \\
    InternVL3.5-8B &all& 76.2 & 33.0 & 51.5 & \phantom{0}8.2 & 27.7 & \phantom{0}0.7 & \phantom{0}9.8 & \phantom{0}0.0 &  100.0\phantom{0} & 54.0 & 15.8 & 27.7 \\
    InternVL3-78B &all& 78.9 & 37.6 & 60.2 & 11.0 & 34.2 & \phantom{0}4.1 & 14.0 & \phantom{0}0.6 &  98.4 & 59.6 & 19.1 & 34.3 \\
    InternVL3.5-38B &all& 79.4 & 40.7 & 56.7 & 12.0 & 27.1 & \phantom{0}0.7 & 10.4 & \phantom{0}0.0 &  100.0\phantom{0} & 57.0 & 20.2 & 34.7 \\
    Claude-3.7 &all& 68.5 & 21.0 & 55.4 & \phantom{0}6.6 & 30.1 & \phantom{0}2.1 & 14.0 & \phantom{0}1.8 &  91.4 & 53.1 & 10.9 & 35.4 \\
    Claude-4.5 &all& 70.1 & 21.0 & 61.2 & 10.8 & 40.4 & \phantom{0}6.5 & 17.1 & \phantom{0}3.0 &  85.7 & 57.6 & 13.3 & 36.4 \\
    Claude-4 &all& 67.9 & 22.5 & 60.6 & 10.7 & 39.7 & \phantom{0}6.2 & 17.7 & \phantom{0}2.4 &  89.3 & 56.6 & 13.8 & 36.6 \\
    GPT-4.1 &all& 70.2 & 22.2 & 61.0 & \phantom{0}9.9 & 40.1 & \phantom{0}5.8 & 18.9 & \phantom{0}4.3 &  80.6 & 57.8 & 13.4 & 36.3 \\
    Qwen2.5-VL-72B &all& 77.5 & 39.5 & 59.5 & 14.1 & 30.5 & \phantom{0}0.7 & 16.5 & \phantom{0}1.2 &  97.0 & 58.4 & 20.6 & 37.3\\
    Qwen3-VL-32B &all& \multicolumn{1}{>{\columncolor{hlgreen}}c}{92.0} & \multicolumn{1}{>{\columncolor{hlgreen}}c}{69.9} & \multicolumn{1}{>{\columncolor{hlgreen}}c}{84.9}& \multicolumn{1}{>{\columncolor{hlgreen}}c}{48.5} & \multicolumn{1}{>{\columncolor{hlgreen}}c}{74.3} & \multicolumn{1}{>{\columncolor{hlgreen}}c}{31.8} & \multicolumn{1}{>{\columncolor{hlgreen}}c}{55.5} & \multicolumn{1}{>{\columncolor{hlgreen}}c}{20.1} & \multicolumn{1}{>{\columncolor{hlgreen}}c}{71.2} & \multicolumn{1}{>{\columncolor{hlgreen}}c}{83.7} & \multicolumn{1}{>{\columncolor{hlgreen}}c}{51.9} & \cellcolor{hlgreen}73.6\\
    \bottomrule
\end{tabular}
}
    \caption{Model's performance on GeoLaux. ACS = Answer Correctness Score, PCS = Process Correctness Score, PQS = Process Quality Score. $\Delta$PCS = ( PCS$_\mathrm{1-4 Steps}$ - PCS$_\mathrm{13-24 Steps}$ ) / PCS$_\mathrm{1-4 Steps}$, measures performance drop as steps increase. The “T*” in the table is an abbreviation for “Thinking” in the full model name.} 
    \label{tab:main_results}
\end{table*}
From solutions generated by o3, Qwen3-VL-32B-Thinking, and GPT-4.1, we randomly sampled 1000 instances and evaluated them using four advanced models. We conducted a manual review to verify the accuracy of these evaluators on both answer and step levels, as summarized in Table \ref{tab:evaluator}.
The results indicate that o4-mini and Gemini-2.5-Pro achieve substantially high evaluation accuracy when aided by reference solutions, validating the reliability of our evaluation framework. 
Prior research \citep{zhang2024geoeval, zhang2024mathverse, wang2025mv} has also demonstrated the evaluator promising potential of advanced MLLMs. 
Comprehensive cases demonstrating the flexibility with alternative approaches and the specific performance of the four evaluators are detailed in Appendix \ref{appendix_evacase}.

\section{Experiments}
\label{sec:Experiments}

Table \ref{tab:models_and_institutions} shows our evaluation covers 23 state-of-the-art MLLMs, with 10 thinking models and 13 non-thinking models. 
Among these, the 11 open-source models are executed on NVIDIA A100 GPUs. All models generate answers through one-shot method. o4-mini is selected as the final evaluator.

Due to prohibitive computational costs of o1 and o3 on the full dataset, we constructed GeoLaux-mini comprising 330 problems uniformly sampled from the original 2186 questions. This subset preserves the original distributions of both step lengths and auxiliary lines to ensure equitable assessment conditions, details in Appendix \ref{appendix_dataset}.  

\subsection{Main Results} 
\begin{figure}[t]
\centering
\includegraphics[width=0.9\columnwidth]{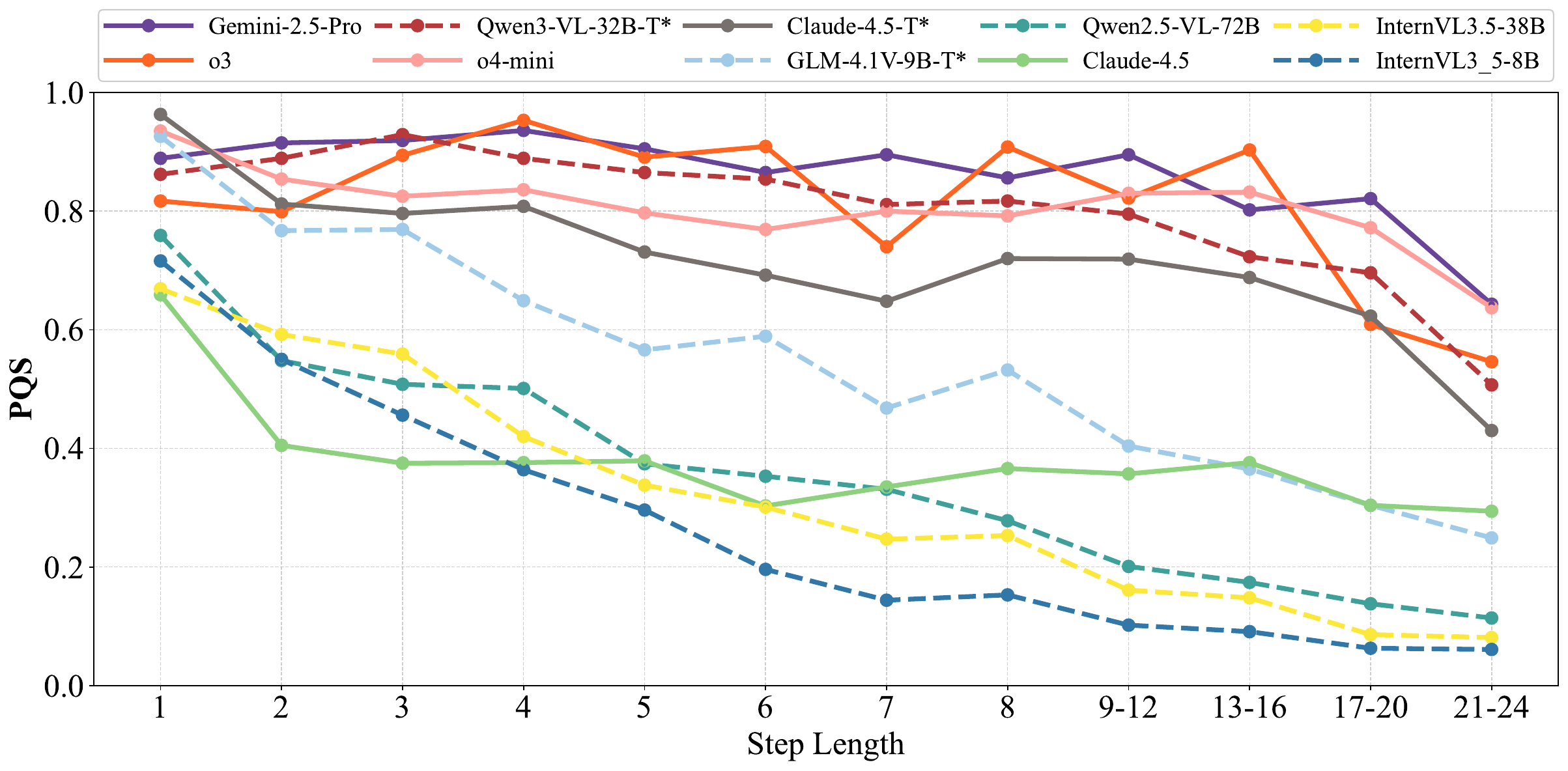} 
\caption{PQS across different step length problems.}
\label{fig:PQS_bystep}
\end{figure}
Table \ref{tab:main_results} presents the performance of 23 MLLMs on GeoLaux across different problem categories (short-step, medium-step, long-step, and ultra-long-step problems), ranked overall by PQS which represents reasoning capability. 
All models exhibit a certain gap between ACS and PCS, indicating the presence of False Positives and underscoring the unreliability of relying solely on answer correctness while ignoring the solution process. Therefore, PCS is adopted as the true measure of correctness.
\paragraph{Models' Ranking.} As demonstrated, thinking models significantly outperform non-thinking ones. Gemini-2.5-Pro, o3, and Qwen3VL-32B-Thinking achieve top three PQS, demonstrating superior reasoning capability. Meanwhile, o3 and o4-mini claim the top spots for PCS and ACS respectively. Among non-thinking models, Qwen3-VL-32B performs the best, leading across all metrics within its category, though a gap remains compared to the strongest thinking models. 
Notably, while current models achieve high ACS, the highest PCS on ultra-long-step problems is only 63.4 (o4-mini), highlighting the necessity of long-step datasets.

\subsection{Long Step Analysis}

"$\Delta$PCS" of Table \ref{tab:main_results} measures performance degradation of models from steps 1-4 to steps 13-24, revealing 18 models exhibit a performance drop exceeding 50\%, with some declines surpassing 90\% or even reaching 100\%. Throughout this transition, o4-mini demonstrates the most stable performance, despite still having a PCS decrease of 19.0\%.

\begin{figure}[t]
\centering
\includegraphics[width=0.9\columnwidth]{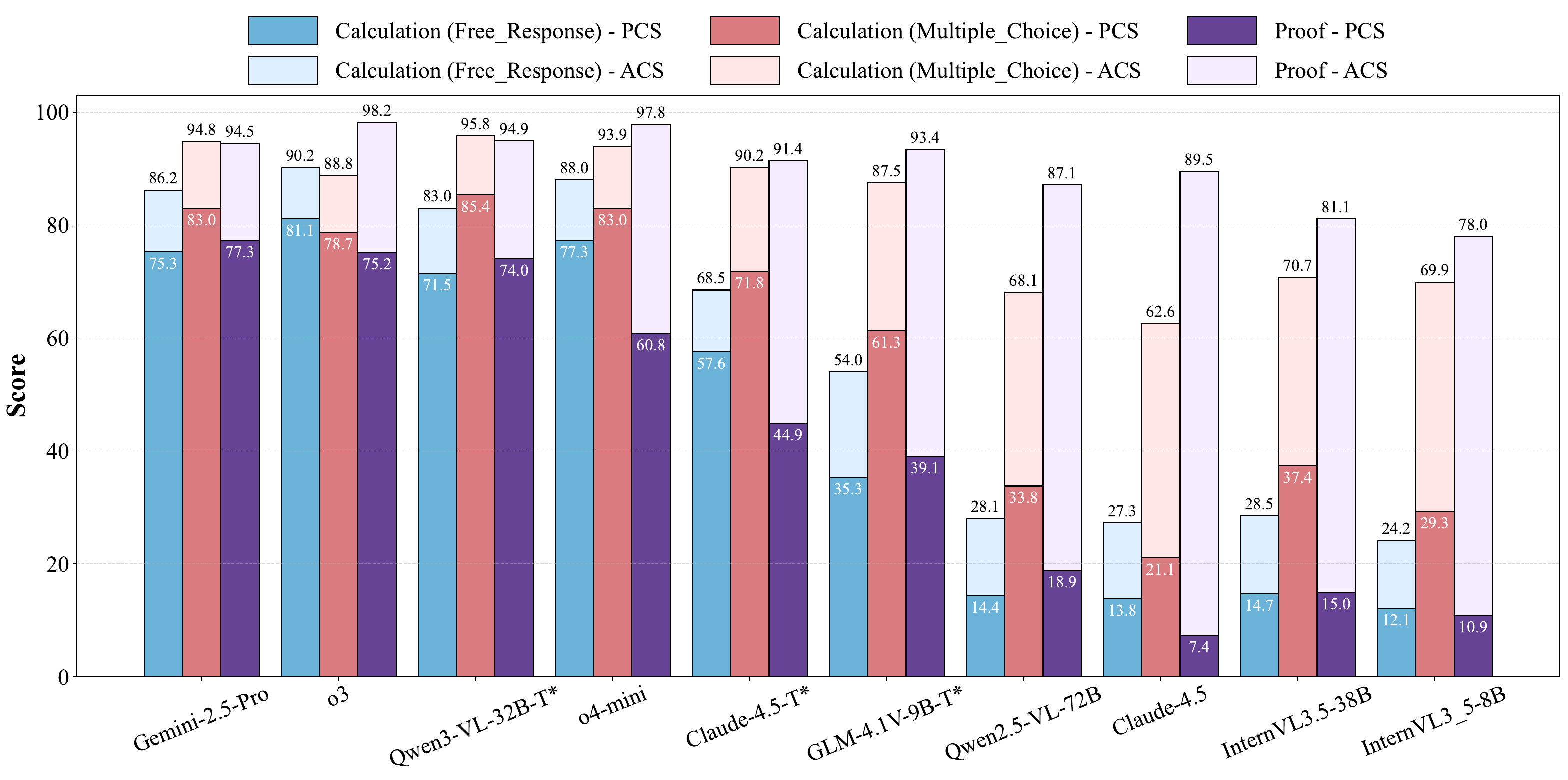} 
\caption{Comparison of ACS and PCS across free-response, multiple-choice, and proof problems.}
\label{fig:Proof_Cal}
\end{figure}
As problem step length increases, we also observe a universal decline in the reasoning quality of all models, as shown by the clear downward trend in PQS (Figure \ref{fig:PQS_bystep}). This trend reflects the increasing difficulty of sustaining precise deduction as inference depth grows.
Meanwhile, most thinking models achieve far higher PQS than non-thinking models, indicating non-thinking models not only make more errors in long-step problems but also tend to make errors at earlier steps of reasoning—since earlier steps carry greater weight in the PQS formula (Equation~\ref{eq:weight_function}). This exposes their relatively weak capability in long-step reasoning.

\subsection{False Positive Analysis}

\label{sec:False Positive Analysis}
In our work, False Positives refer to cases where answer is correct but process contains errors, leading to discrepancies between ACS and PCS. 
We analyzed the distribution of False Positives across problem types. As shown in Figure \ref{fig:Proof_Cal}, this issue is far more severe in multiple-choice and proof problems than in free-response problems, attributed to chance guessing in multiple-choice tasks and the direct provision of target conclusions in proof tasks.

\textbf{Limited hints can improve model performance, but clear answers make them lazy.}
Figure \ref{fig:Proof_Cal} also reveals an intriguing phenomenon:  nearly all models' PCS exhibit a substantial increase for multiple-choice questions, yet suffer a precipitate decline in proof problems. This suggests that options may act as an incentive or a guide, assisting models in solving problems diligently to reach correct answers. However, when presented with correct answers directly, models become less rigorous about the solving process and resort to shortcuts (as cases in Appendix \ref{appendix_evacase}). They pair correct final answers with flawed reasoning, essentially "deceiving" the users. This phenomenon warrants careful attention in future research.

\subsection{Auxiliary Line Analysis}
\begin{figure}[t]
\centering
\includegraphics[width=0.9\columnwidth]{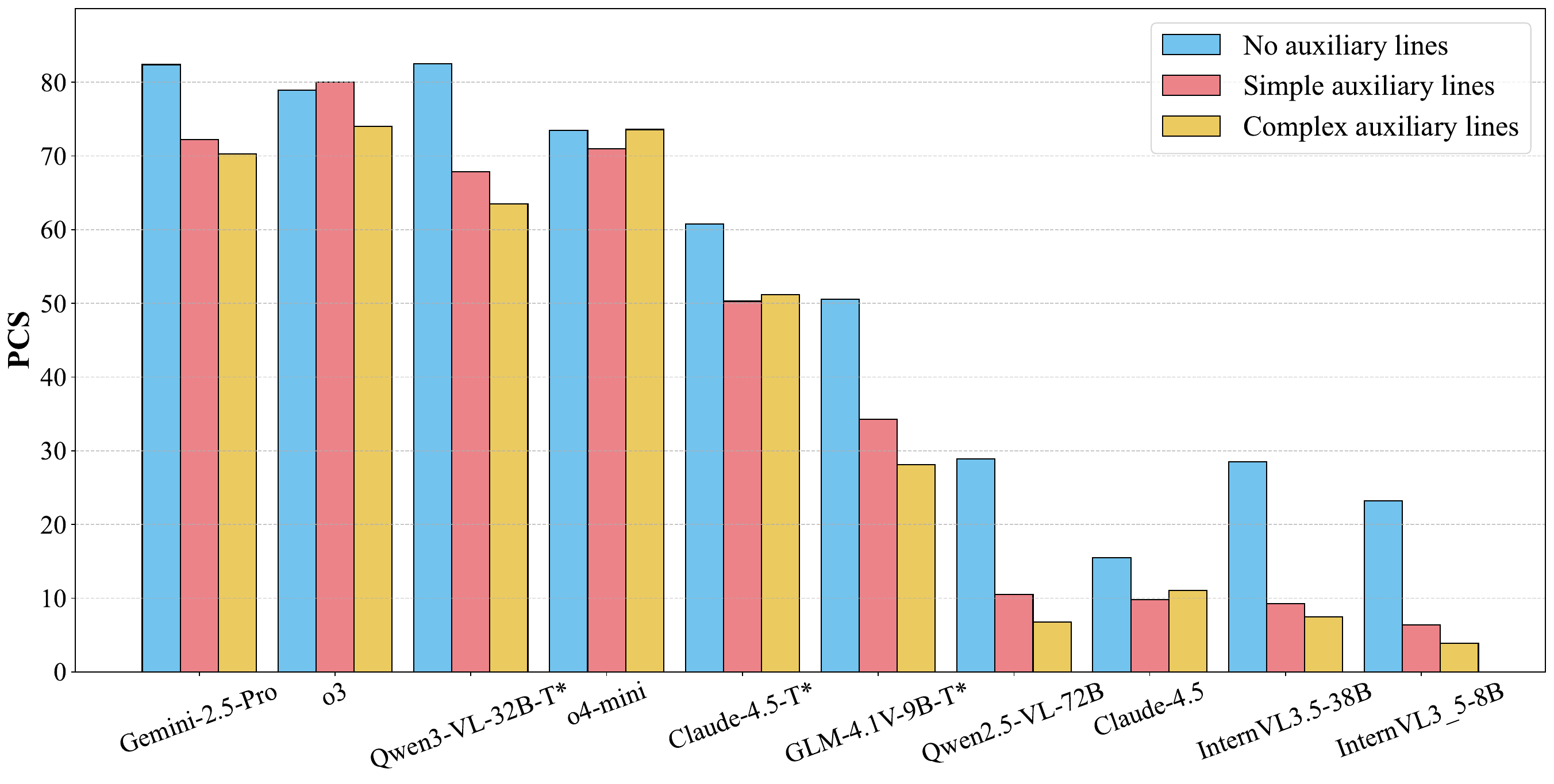} 
\caption{PCS under different auxiliary line complexity. }
\label{fig:Auxiliary_TPS}
\end{figure}

\begin{table}[t]
    \centering
    \setlength{\tabcolsep}{1mm}
    \scalebox{0.75}{\begin{tabular}{@{} l | c c@{}}
        \toprule
        \textbf{Model} & \textbf{Construction Rate(\%)} & \textbf{Construction PCS}\\
        \midrule
        Claude-4.5-T* & \phantom{0}6.4 / 56.4 & 33.3 / 44.3  \\
        o3 & 10.0 / 41.4 & 57.1 / 72.4 \\
        Gemini-2.5-Pro & \multicolumn{1}{>{\columncolor{hlblue}}c}{28.6 / 83.6} & \multicolumn{1}{>{\columncolor{hlblue}}c}{72.5 / 82.1}\\
        \midrule
        GPT-4.1 & 15.0 / 17.9 &\phantom{0}4.7 /\phantom{0}8.0\\
        Qwen3-VL-32B & \multicolumn{1}{>{\columncolor{hlgreen}}c}{38.6 / 56.4} & \multicolumn{1}{>{\columncolor{hlgreen}}c}{35.6 / 44.3} \\
        \bottomrule
    \end{tabular}}
    \caption{Comparison of self-attempt and heuristic prompting across construction rate and PCS.}
    \label{tab:heuristic_results}
\end{table}

As shown in \figurename~\ref{fig:Auxiliary_TPS}, problems requiring auxiliary lines (especially complex ones) significantly challenge MLLMs, causing a noticeable drop in PCS compared to those without. 
o3 and o4-mini do not show a significant decline, because they often resort to coordinate-based methods when facing problems requiring auxiliary lines. Unfortunately, this approach escalates computational costs and solution complexity, undermining human interpretability.

Following Section \ref{sec:Meta Auxiliary}, we compare the auxiliary line construction rates and the resulting PCS between autonomous solving and heuristic prompting, as summarized in Table \ref{tab:heuristic_results}. 
For both "Construction Rate (\%)" and "Construction PCS", values to the left and right of "/" correspond to autonomous solving and heuristic auxiliary prompting, respectively.

This performance comparison reveals two key limitations: \textbf{(1) Weak awareness of autonomously constructing auxiliary lines:} The rate of MLLMs autonomously constructing auxiliary lines is extremely low. This sharply contrasts with the high construction rate when heuristically prompted with the correct lines, indicating that models often simply fail to realize they should try to construct an auxiliary line. \textbf{(2) Limited capability in generating correct auxiliary lines:} For problems where models proactively constructed auxiliary lines, their resulting PCS remains low and is significantly outpaced by the PCS achieved when using prompted standard auxiliary lines. This demonstrates that even when models do attempt to draw auxiliary lines, they frequently struggle to generate the most appropriate and mathematically correct auxiliary constructions to facilitate the solution.

Interestingly, Table \ref{tab:heuristic_results} reveals that Non-Thinking MLLMs have a generally higher autonomously construction rate compared to Thinking models, but achieve a lower PCS. This divergence stems from differing reasoning strategies. Thinking models (e.g., Gemini-2.5-Pro) evaluate multiple paths and avoid unnecessary constructions that needlessly expand the search space if existing relationships suffice. Conversely, non-thinking models like Qwen3-VL-32B-Instruct tend to rely on mechanical rule application. They exhibit a heuristic tendency to attempt, frequently and blindly applying auxiliary lines to familiar geometric setups without verifying their strategic utility. Consequently, Qwen3-VL-32B-Instruct records the highest proactive construction rate but a substantially lower PCS.

\begin{figure}[t]
\centering
\includegraphics[width=0.9\columnwidth]{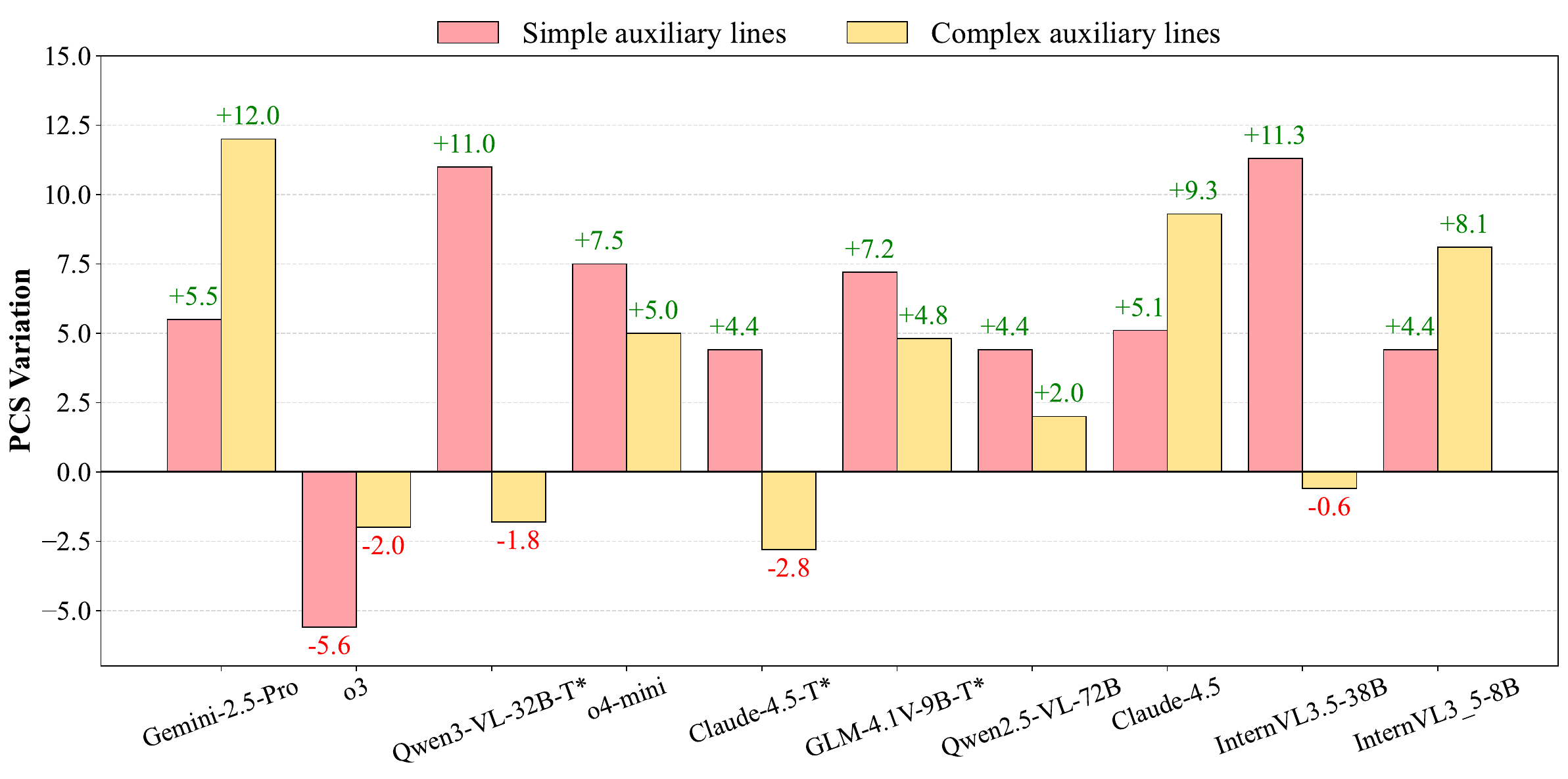} 
\caption{PCS delta after prompting auxiliary lines. }
\label{fig:change_Auxiliary}
\end{figure}
While Table \ref{tab:heuristic_results} evaluates PCS solely on problems where auxiliary lines are constructed, Figure \ref{fig:change_Auxiliary} assesses PCS variations across all prompted instances, irrespective of adoption. Although heuristic prompting generally improves PCS across most models, it reveals critical edge cases. For instance, o3 exhibits a performance drop because when suggested auxiliary lines are not adopted, the explicit hints likely disrupt its inherent coordinate-based strategies. Similarly, a few models benefiting from simple hints falter with complex ones, as they fail to comprehend complex constructions and are overwhelmed by the increased visual complexity. These negative impacts prove that relying solely on external auxiliary lines prompts is insufficient. Therefore, \textbf{advancing models' capabilities to autonomously construct optimal auxiliary lines} must be a primary focus for future research.

\subsection{Error Type Analysis}
\begin{figure}[t]
\centering
\includegraphics[width=0.9\columnwidth]{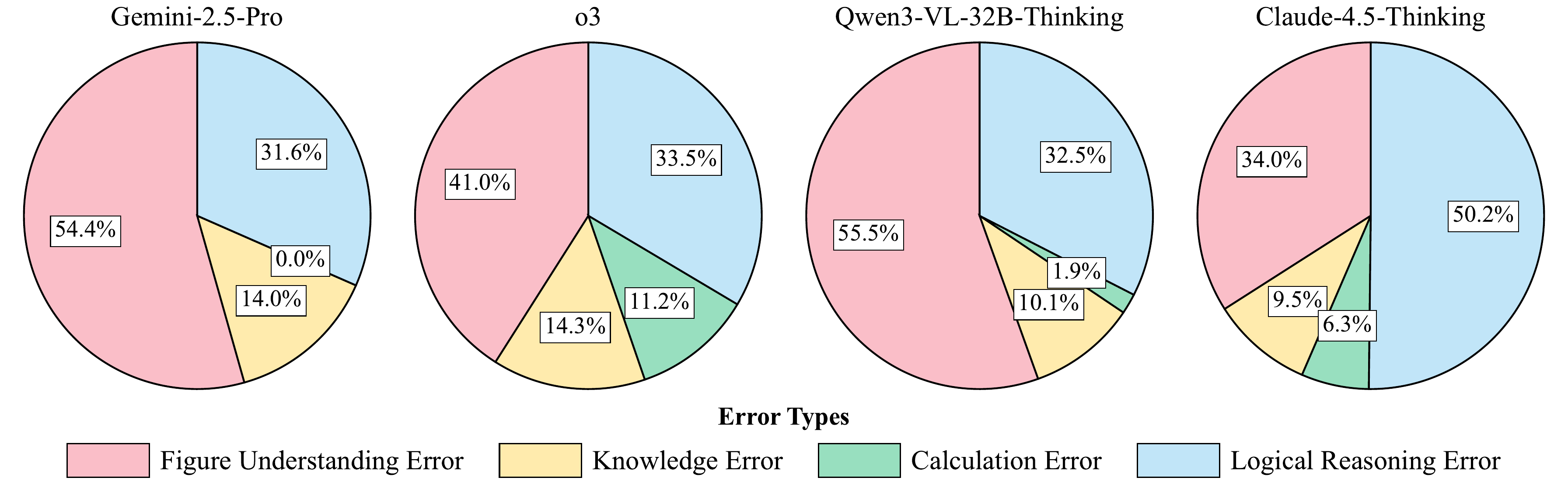} 
\caption{Error types distribution for MLLMs.}
\label{fig:error_pies}
\end{figure}
We analyze the error causes according to the error types defined in section 4.4, and present representative error type distributions in Figure \ref{fig:error_pies}.
The results demonstrate that incorrect geometric figure comprehension and flawed logical reasoning remain the fundamental bottlenecks limiting MLLMs' GPS capabilities. While knowledge errors and calculation errors persist across most models' solutions, these two error types prove relatively more addressable through external tools. Notably, Gemini-2.5-Pro exhibits virtually no calculation error during problem-solving, which likely contributes to its top performance on our benchmark. Appendix \ref{appendix:error_type} contains some error analysis cases.

\section{Conclusion}

In this work, we present GeoLaux, a comprehensive geometric dataset with long-step problems and auxiliary line annotations. Based on this dataset, we evaluate 23 leading MLLMs through a five-dimensional framework, revealing: (1) a severe performance drop in long-step reasoning, (2) limited hints can improve model performance, and (3) the pivotal role of auxiliary line construction in GPS. These insights offer valuable guidance for advancing future MLLMs’ geometric reasoning.

\section*{Acknowledgements}
This work was supported by Fundamental and Interdisciplinary Disciplines Breakthrough Plan of the Ministry of Education of China (JYB2025XDXM116), National Natural Science Foundation of China (No. 62137002, 62293550, 62293553, 62293554, 62450005, 62437002, 62477036, 62477037, 62577043, 62192781),  the Shaanxi Provincial Social Science Foundation Project (No. 2024P041), the Youth Innovation Team of Shaanxi Universities "Multi-modal Data Mining and Fusion".

\section*{Limitations}

While GeoLaux provides a robust benchmark for evaluating geometric reasoning, three key limitations warrant discussion. First, our dataset exclusively features problems at the Chinese High School Entrance Examination (Zhongkao) level, as we prioritize investigating foundational reasoning flaws relevant to real-world education. Although GeoLaux already poses a substantial challenge to existing models, incorporating higher-difficulty, Olympiad-level problems remains a direction for future research as models advance. Second, our evaluation framework relies on MLLM evaluator. While o4-mini demonstrates superior accuracy, employing it imposes substantial computational and financial costs when evaluating the full set of 2186 problems. This may constrain the feasibility of frequent, large-scale iterative testing. Finally, our current method for analyzing the impact of auxiliary lines involves heuristically prompting models with standard, ground-truth auxiliary lines. This approach simplifies the process and differs from realistic scenarios where models must autonomously construct auxiliary lines—a complex capability that remains distinct from merely utilizing provided hints. In future work, we will focus on optimizing evaluation efficiency and investigating the autonomous generation of geometric constructions to better reflect real-world problem-solving scenarios.

\section*{Ethical Considerations}
\label{appendix_Ethical}
In constructing GeoLaux, we strictly adhered to ethical guidelines and best practices. To ensure high-quality annotation, we employed two Master's students, providing them with comprehensive operational guidelines and offering fair compensation for their contributions. The dataset sources problems exclusively from publicly available examination materials, which subsequently underwent rigorous cleaning, deduplication, and standardization to guarantee data reliability while minimizing potential biases. Furthermore, we confirm that the dataset does not contain any personally identifiable information (PII) or private content. We are also committed to preventing data contamination; as detailed in Appendix \ref{appendix_Leakage}, we conducted extensive decontamination checks to ensure the validity of our evaluation. During writing, AI assistants were utilized solely for linguistic polishing to enhance readability. To promote open research while respecting copyright, we release the dataset and associated scripts under MIT and CC BY-NC-SA 4.0 licenses, strictly prohibiting commercial use.

\nocite{team2023gemini} \nocite{liu2023visual}
\nocite{jiang2025corvid}

\nocite{Zhang_Wang_Zhu_Cheng_He_Li_Lin_Liu_Cambria_2026,Zhang_Wang_Wang_Xu_Lin_Zhang_Mao_Cambria_Liu_2026,zhang2026geochallenge}

\bibliography{custom}
\newpage
\section*{Appendix Overview} 

\begin{itemize}
    \item Section \ref{appendix_dataset}: GeoLaux Details.
    \item Section \ref{appendix_pqs}: PQS discussions. 
    \item Section \ref{appendix_promptsansmodel}: Prompts and Model Details.
    \item Section \ref{appendix_evacase}: Process Evaluation Cases.
    \item Section \ref{appendix:error_type}: Error Type Cases.
    \item Section \ref{appendix_Leakage}: Data Leakage Analysis.
\end{itemize}
\appendix

\section{GeoLaux Details} 
\label{appendix_dataset}
\subsection{GeoLaux-mini Details}
\label{appendix_minidataset}
We performed uniform sampling on GeoLaux to create GeoLaux-mini, a 330-problem subset specifically designed for testing computationally expensive models (o1, o3) and conducting supplementary auxiliary line heuristic evaluations.
\begin{figure}[h]
\centering
\includegraphics[width=1\columnwidth]{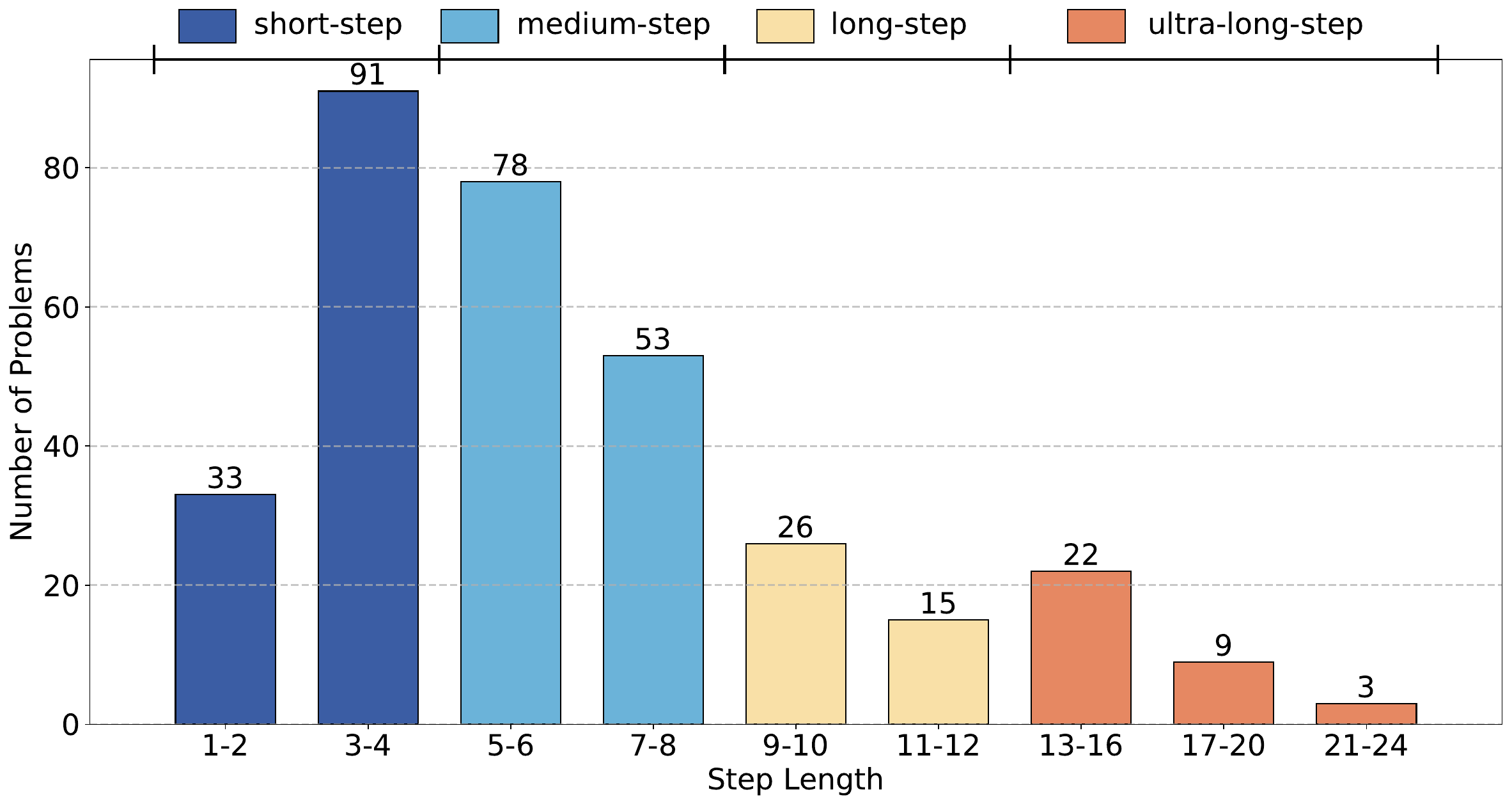} 
\caption{Problem quantity statistics across step lengths in GeoLaux-mini. }
\label{fig:Step_Number_count_mini}
\end{figure}

\begin{figure}[h]
\centering
\includegraphics[width=0.35\columnwidth]{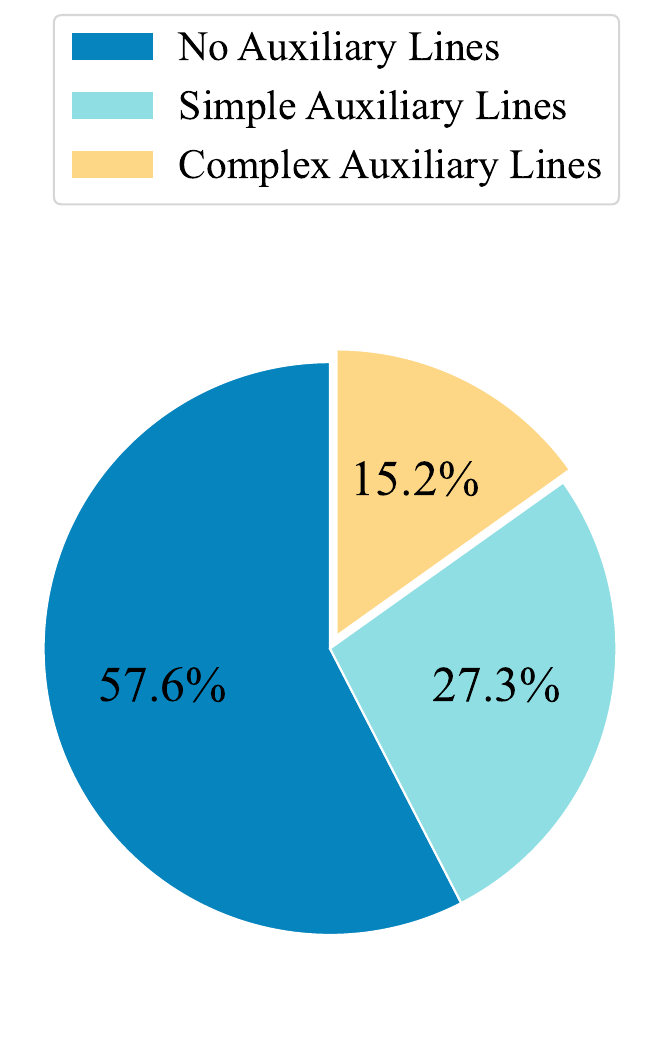} 
\caption{Distribution of auxiliary line types in GeoLaux-mini. }
\label{fig:Auxiliary_distribution_mini}
\end{figure}

GeoLaux-mini maintains a similar step-length distribution to the original dataset (as Figure \ref{fig:Step_Number_count_mini}), containing a substantial number of medium-step, long-step, and ultra-long-step problems. The subset comprises 109 proof problems and 221 calculation problems, 190 problems that do not require auxiliary line and 140 problems that need auxiliary line. Its auxiliary line distribution illustrated in Figure \ref{fig:Auxiliary_distribution_mini}.

\subsection{Language robustness analysis}
Due to the data source, the original language of the dataset is Chinese. However, the leading models we tested are inherently multilingual, having been pre-trained on massive Chinese corpora to achieve native-level reading comprehension. Additionally, middle school geometry problems feature highly concise text, standardized vocabulary, and formulaic syntactic structures (e.g., "As shown in the figure, given A, prove B"). As a result, the linguistic processing barrier is exceptionally low, minimizing the risk of tokenization artifacts. Therefore, the simple terminology used in these problems does not pose a meaningful challenge to the models' comprehension.

To further validate the linguistic robustness of our dataset, we translated the 330 problems from GeoLaux-mini into English (the translated dataset is also open-sourced on GitHub) and evaluated five representative large models. The performance of these models on GeoLaux-mini before and after translation is presented in Table \ref{tab:language_robustness}.
\begin{table*}[t]
    \centering
    \setlength{\tabcolsep}{1mm}
    \scalebox{0.7}{\begin{tabular}{@{} l | c c c c c c c@{}}
        \toprule
        \textbf{Model} & \textbf{1-4 Steps PCS} & \textbf{5-8 Steps PCS} & \textbf{9-12 Steps PCS} & \textbf{13-24 Steps PCS} & \textbf{AVG ACS} & \textbf{AVG PCS} & \textbf{AVG PQS}\\
        \midrule
        Claude-4.5-T* & 71.0 / 68.1 (-2.9) & 49.8 / 47.3 (-2.5) & 46.5 / 48.0 (+1.5) & 32.1 / 29.1 (-3.0) & 81.2 / 81.2 (+0.0) & 58.6 / 56.1 (-2.5) & 74.7 / 73.8 (-0.9)\\
        o3 & 83.9 / 83.1 (-0.8) & 80.9 / 84.0 (+3.1) & 73.2 / 75.9 (+2.7) & 53.6 / 57.1 (+3.5) & 92.4 / 89.1 (-3.3) & 78.5 / 77.8 (-0.7) & 86.0 / 84.9 (-1.1)\\
        Gemini-2.5-Pro & 85.5 / 87.1 (-1.6) & 80.4 / 79.4 (-1.0) & 73.0 / 75.7 (+2.7) & 46.3 / 50.0 (+3.7) & 89.4 / 85.8 (-3.6) & 77.6 / 78.8 (+1.2) & 88.8 / 89.9 (+1.1)\\
    \midrule
        GPT-4.1 & 25.8 / 26.2 (+0.4) & \phantom{0}9.2 / 12.5 (+3.3) & \phantom{0}9.8 / \phantom{0}7.3 (-2.5) & \phantom{0}5.3 / \phantom{0}7.1 (+1.8) & 55.8 / 59.7 (+3.9) & 14.5 / 16.4 (+1.9) & 35.1 / 37.3 (+2.2)\\
        Qwen3-VL-32B & 77.4 / 74.2 (-3.2) & 53.4 / 48.9 (-4.5) & 35.3 / 39.9 (+4.6) & 23.9 / 25.7 (+1.8) & 85.2 / 86.1 (+0.9) & 55.8 / 56.4 (+0.6) & 73.3 / 74.2 (+0.9)\\
    \bottomrule
    \end{tabular}}
    \caption{Model performance on the GeoLaux dataset in Chinese (left of /) and English (right of /). Values in "()" represent the change in scores when transitioning from Chinese to English problems.}
    \label{tab:language_robustness}
\end{table*}

As observed, the performance variations among different models before and after translation are inconsistent—some increase while others decrease—and the magnitude of these changes is consistently small, representing normal random fluctuation. This demonstrates that our experimental results are fundamentally unaffected by the language used, confirming their robustness.
\subsection{GeoLaux Examples}
\begin{figure*}[t!] 
\centering

\begin{minipage}[t]{0.48\textwidth}
    \centering
    \includegraphics[width=\textwidth]{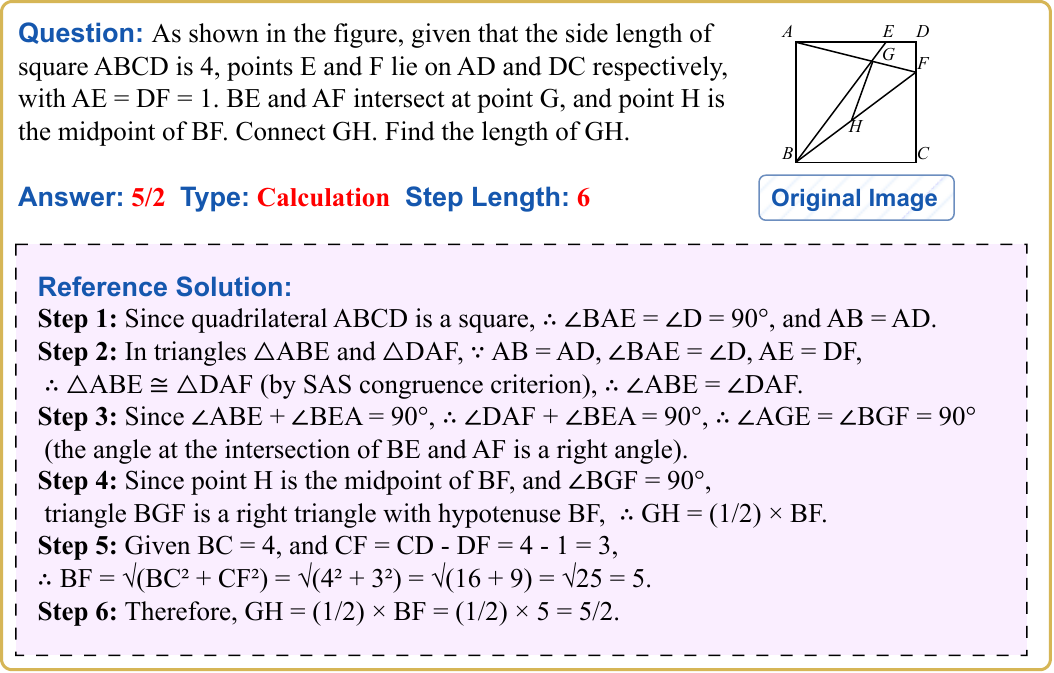}
    \vspace{0.5em} 
    {\small \textbf{(a)} Calculation problem}
    \label{fig:appendix_c_1}
\end{minipage}
\hfill
\begin{minipage}[t]{0.48\textwidth}
    \centering
    \includegraphics[width=\textwidth]{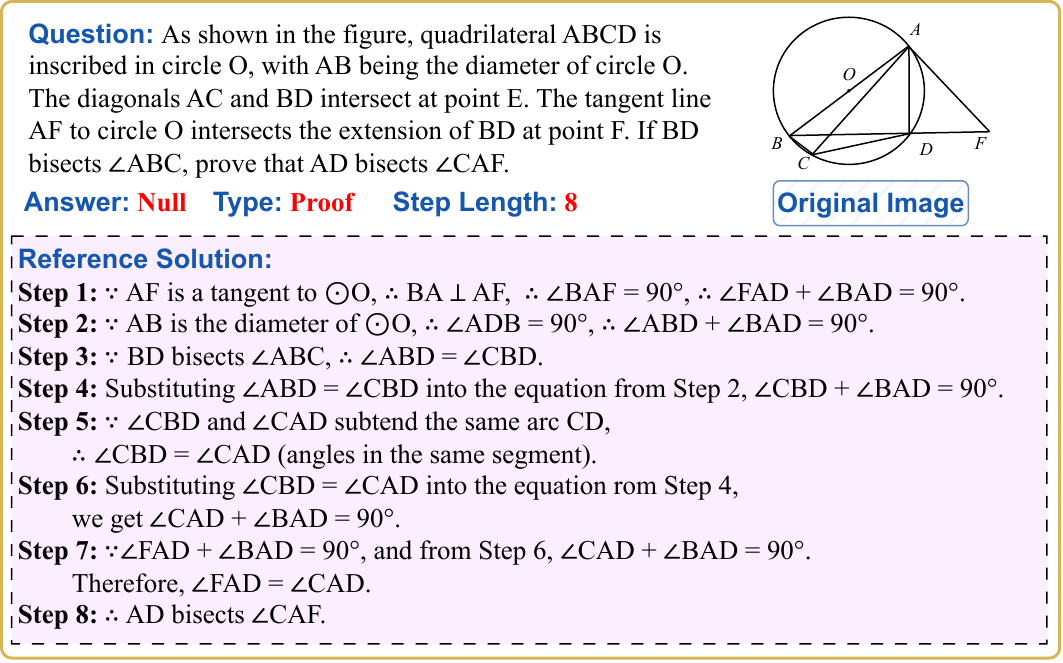}
    \vspace{0.5em}
    {\small \textbf{(b)} Proof problem}
    \label{fig:appendix_c_proof}
\end{minipage}

\vspace{0.5em} 

\begin{minipage}[t]{0.48\textwidth}
    \centering
    \includegraphics[width=\textwidth]{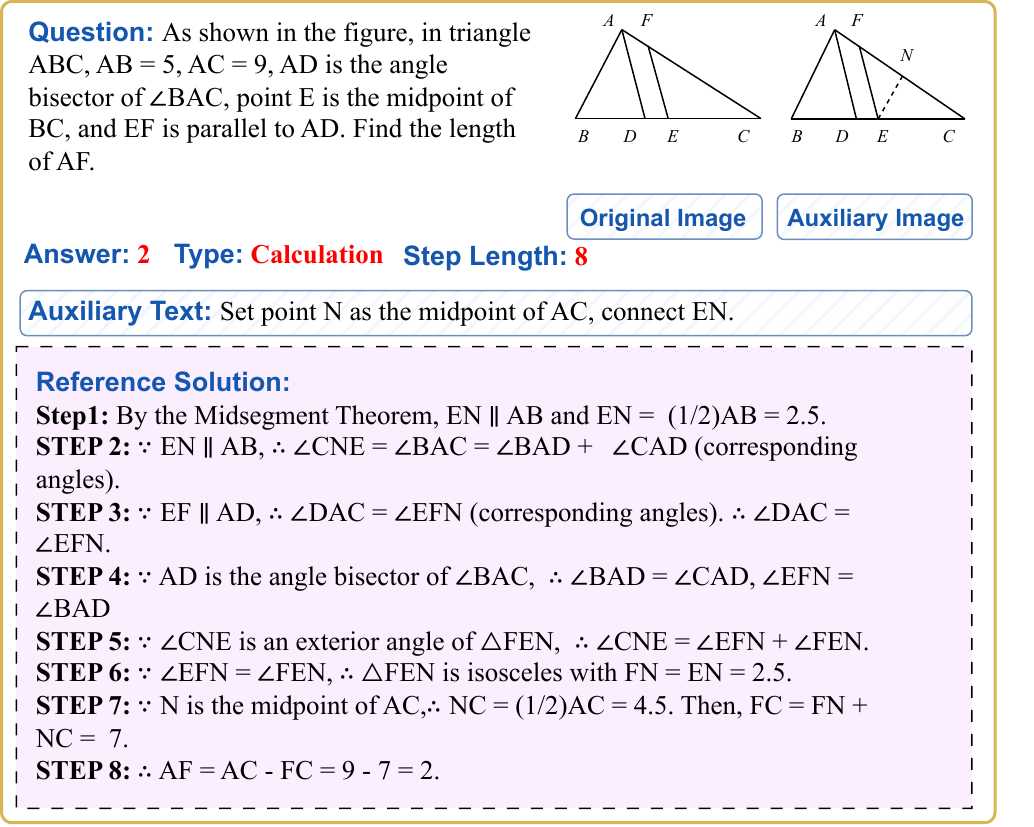}
    \vspace{0.5em}
    {\small \textbf{(c)} Auxiliary lines construction problem}
    \label{fig:appendix_c_calc_1}
\end{minipage}
\hfill
\begin{minipage}[t]{0.48\textwidth}
    \centering
    \includegraphics[width=\textwidth]{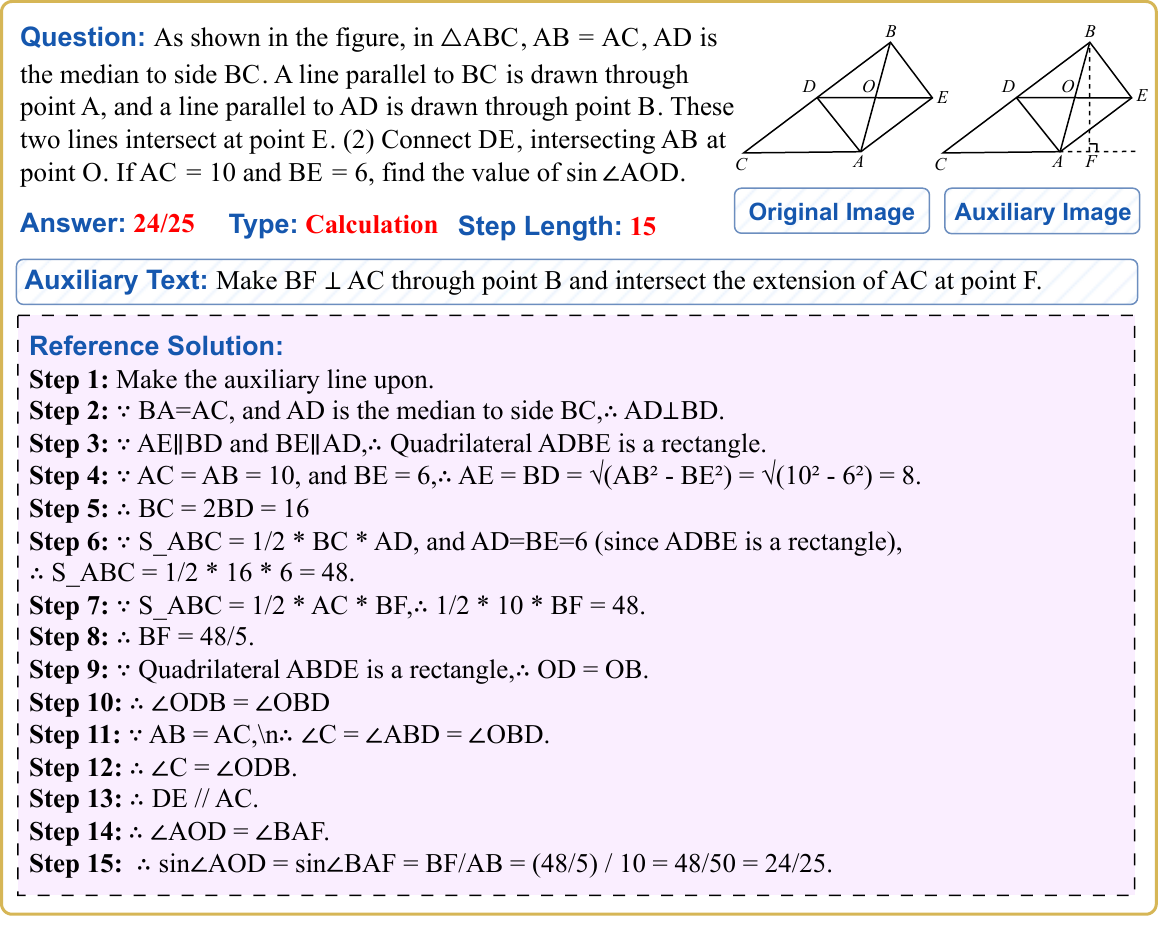}
    \vspace{0.5em}
    {\small \textbf{(d)} Ultra-long step problem}
    \label{fig:appendix_c_calc_2}
\end{minipage}

\caption{Examples from the GeoLaux dataset.}
\label{fig:dataset_example}
\end{figure*}

The GeoLaux dataset encompasses a comprehensive collection of geometry problems that can be classified along three key dimensions: (1) by the presence of solvable answers in the questions, differentiating between calculation problems and proof problems; (2) by the necessity of auxiliary construction, distinguishing problems requiring auxiliary lines from those needing none; and (3) by solution step length, categorizing problems as short-step, medium-step, long-step, or ultra-long-step problems. Representative examples  are illustrated in Figure~\ref{fig:dataset_example}.

\section{PQS discussions} 
\label{appendix_pqs}
PQS score comprises two components: the step weight function and the score activation function. For an MLLM-generated solution process, evaluator scores each individual reasoning step, assigning 1 for correct steps and 0 for incorrect ones. Step weight function assigns specific weights to each step's score, then score activation function ultimately computing a weighted overall process quality score. These two functions are discussed separately below.
\subsection{Step Weight Function}
The design rationale for the step weight function is guided by the following principles:

\textbf{(1) Decreasing function.} In long-step problems, the later an error occurs, the stronger we consider the model's long-step reasoning capability to be. Consequently, models failing in the initial steps should incur heavier penalties, resulting in lower final scores. This logic implies that higher weights should be assigned to earlier steps, suggesting that the step weight function should be monotonically decreasing.

\begin{figure}[t]
    \centering
    \begin{minipage}{0.36\textwidth} 
        \centering
        \includegraphics[width=\textwidth]{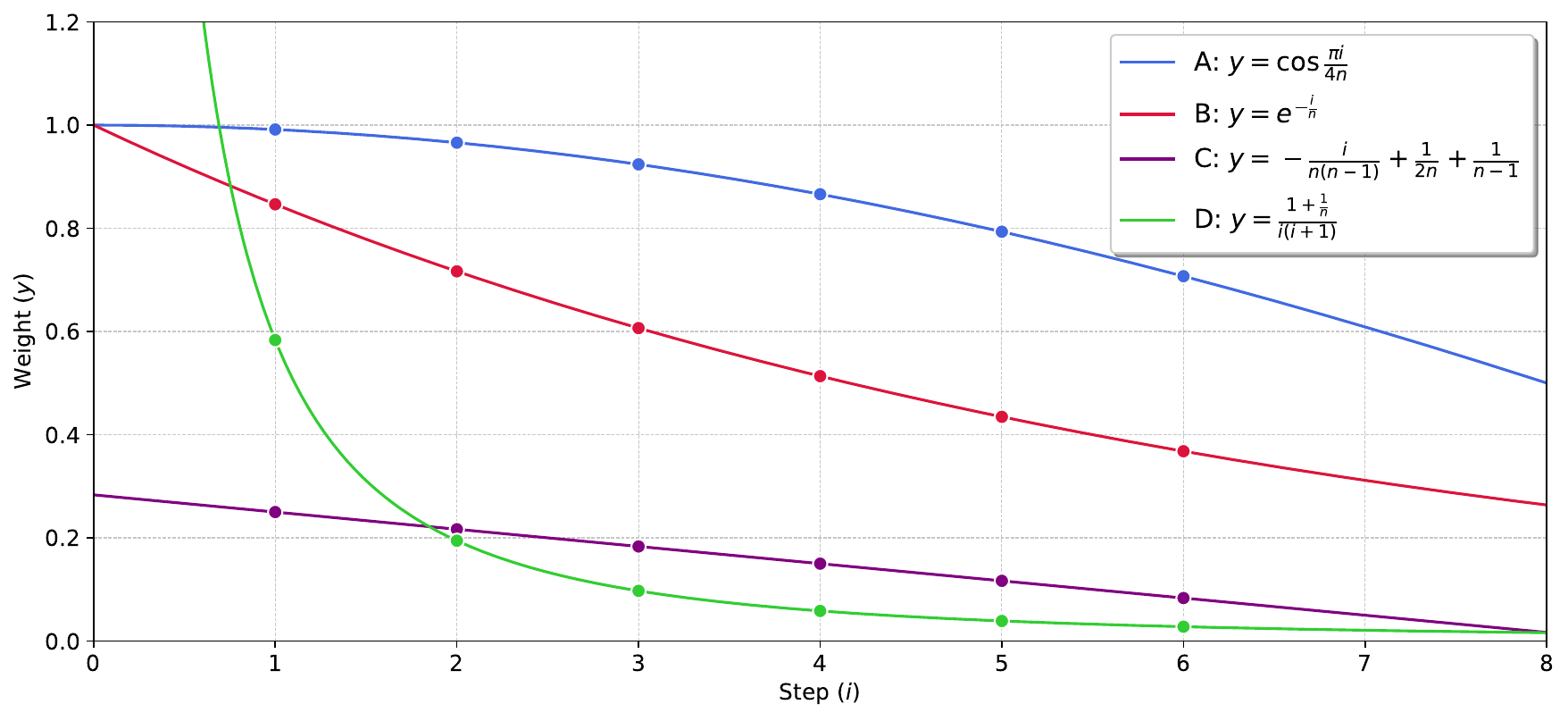}
        {\small (a) Contrast of various Step Weight Functions, illustrated with 6-step reasoning process ($n=6$).}
        \label{fig:weight_function}
    \end{minipage}
    \hspace{-0.2cm}
    \begin{minipage}{0.115\textwidth} 
        \centering
        \includegraphics[width=\textwidth]{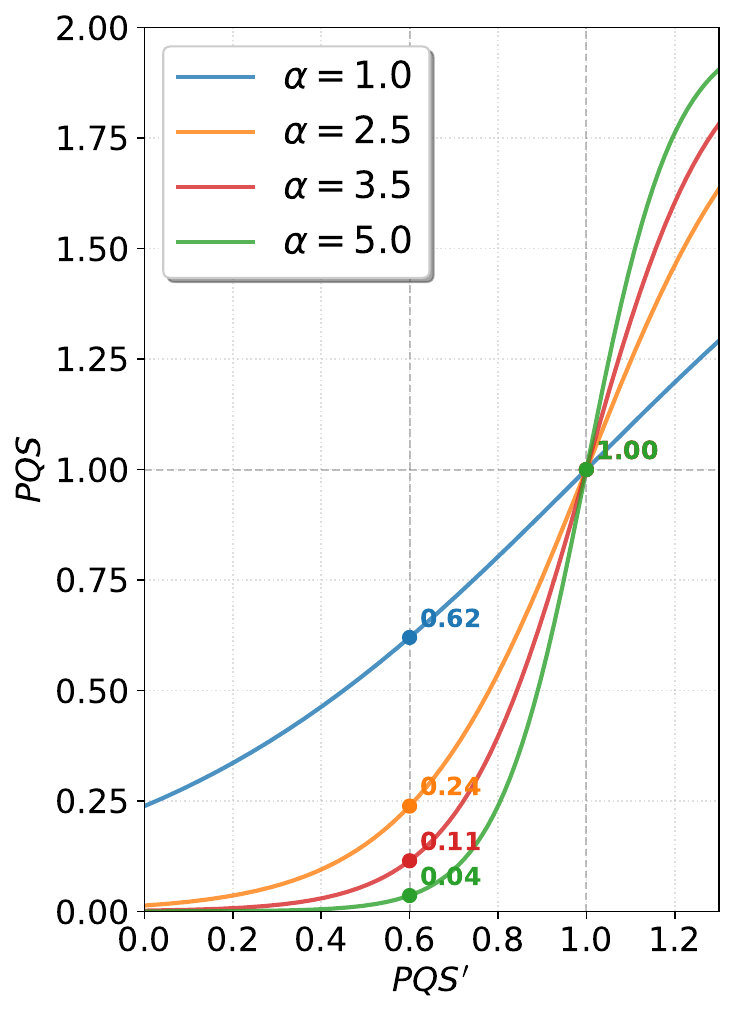}
        {\small (b) Contrast of different $\alpha$ in PQS}
        \label{fig:right}
    \end{minipage}
    
    \caption{Diagram of the functional analysis in PQS.}
    \label{fig:pqs_discussion}
\end{figure}

We propose several decreasing weight functions; their functional trends and the specific weights assigned to each step (for $n=6$) are illustrated in Figure \ref{fig:pqs_discussion} (a). The functions shown in the figure are: the blue Function A:
\[
y = \cos \frac{\pi i}{4n};
\] 
the red Function B: 
\[y = e^{-\frac{i}{n}};\]
the purple Function C: \[y = -\frac{i}{n(n-1)} + \frac{1}{2n} + \frac{1}{n-1};\] 
and the green Function D: \[y = \frac{1 + \frac{1}{n}}{i(i+1)}.\]All are monotonically decreasing and together cover various types of functions—linear, concave, and convex.

\textbf{(2) Convex function:} The importance gap is larger for earlier steps and smaller for later ones. For example, if two models make mistakes at step 2 and step 4 respectively, it indicates that the difference in their capabilities is significantly greater compared to two models making errors at step 12 and step 14. Therefore, the weight function should be a convex function, meaning that the rate of decrease is faster in the early stage than in the later stage.

Based on this consideration, blue function A (concave function) and purple function C (linear function) are excluded.

\textbf{(3) Moderate decreasing rate:} The weighting function should not decrease too rapidly. For long-step problems, the performance in later steps remains critical for evaluation and should retain significant weight.

The green function D is a typical example of an excessively rapid decline. As can be seen, the weights at steps 5 and 6 are extremely small compared to weights at steps 1 and 2. Additionally, the difference between the weights at step 5 and step 6 is minimal, making it unsuitable as a weight function. On the other hand, the red function B, which corresponds to Equation~\ref{eq:weight_function} in the main text, perfectly satisfies all the requirements we have proposed and is therefore selected as the step weight function.

\subsection{Score Activation Function}
By multiplying the score assigned by the evaluator to each step with the corresponding weight of that step, summing up the results, and then normalizing by dividing by the sum of the weights across all steps, the initial process quality score can be obtained. The formula is as follows (which is Equation~\ref{equation_initial_process_score} in the main text):
\[
\mathrm{PQS'} = \sum_{i=1}^n \frac{\eta_i \cdot y_i}{\sum_{j=0}^n y_j} 
  = \sum_{i=1}^n \frac{\eta_i \cdot e^{-\frac{i}{n}} }{\sum_{j=1}^n e^{-\frac{j}{n}}}.
\]

For example, if a solution has 3 steps with scores [1,0,1], the initial process score is 
\[
\begin{aligned}
\mathrm{PQS'}_{\text{\small example}} 
    &= \frac{1\cdot e^{-\frac{1}{3}} + 0\cdot e^{-\frac{2}{3}} + 1\cdot e^{-\frac{3}{3}}}{e^{-\frac{1}{3}}+e^{-\frac{2}{3}}+e^{-\frac{3}{3}}} \\
    &\approx 0.6832
\end{aligned}
\]
However, we observed that since model solutions always contain some correct steps, $\mathrm{PQS'}$ consistently falls between 0.6 and 1, failing to highlight the differences in models’ reasoning capabilities. To address this, we applied the $\tanh$ activation function to $\mathrm{PQS'}$.
\[
\mathrm{PQS} = \tanh\bigl(\alpha (\mathrm{PQS'} - 1)\bigr) + 1,
\]

Using the tanh function for activation, unlike a linear function, can relatively amplify the differences between higher-scoring models to a greater extent, facilitating the selection of the best-performing model.

$\alpha$ is a hyperparameter, and the impact of its different values on the mapping of $\mathrm{PQS}$ is illustrated in Figure \ref{fig:pqs_discussion} (b) (focus on the $\mathrm{PQS}$ value range of 0.6–1). When $\alpha$ is set to 1, the mapped $\mathrm{PQS}$ values still fall within [0.61,1], failing to amplify the differences between models. When $\alpha$ is set to 2.5, the mapped range remains relatively narrow [0.24,1]. On the other hand, when $\alpha$ is set to 5, although it maps $\mathrm{PQS'}$ to a broader range, it significantly reduces the distinctions among lower-performing models, which is detrimental to overall ranking. Therefore, after balancing the objectives of selecting the optimal model and maintaining a comprehensive ranking of all 23 models, we chose $\alpha$ = 3.5 for subsequent experiments. At this value, $\mathrm{PQS}$ is mapped to the interval [0.11,1].

Therefore, for the solution with 3 steps scored as [1,0,1], the final activated PQS value is 
\[
\begin{aligned}
\mathrm{PQS}_{\text{\small example}} 
    &\approx \tanh\bigl(3.5 \cdot (0.6832 - 1)\bigr) + 1 \\
    &\approx 0.3399
\end{aligned}
\]


\section{Prompts and Model Details} 
\label{appendix_promptsansmodel}
\subsection{Prompt for Initial Solution Generation.}
\label{appendix_prompt}
\begin{figure*}[t!]  
\centering
\includegraphics[width=0.9\textwidth]{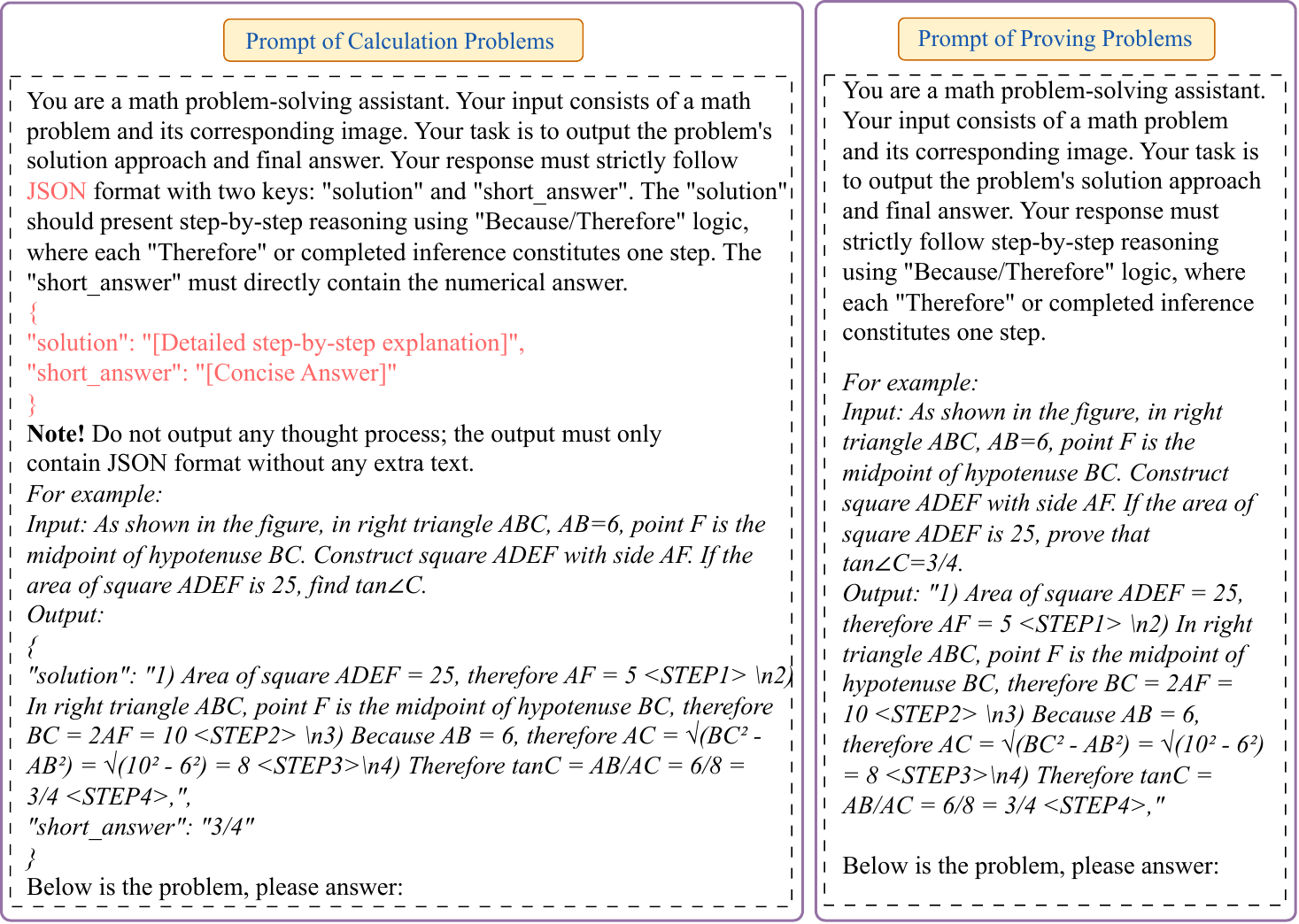} 
\caption{One-shot solution generation prompt for main evaluation.}
\label{fig:prompt_for_maingen}
\end{figure*}

In the main experimental section, we employ one-shot prompt to guide MLLMs in generating responses in JSON format. The use of one-shot prompt ensure all models strictly adhere to our specified JSON format, thereby simultaneously obtaining both the step-by-step solution process (to facilitate subsequent evaluation) and numerical answers for calculation problems. Sample prompts for calculation problems and proof problems are shown in the Figure \ref{fig:prompt_for_maingen}.
\subsection{Prompt for Auxiliary Line Heuristic Solution Generation.}
\begin{figure*}[t!]   
\centering
\includegraphics[width=0.9\textwidth]{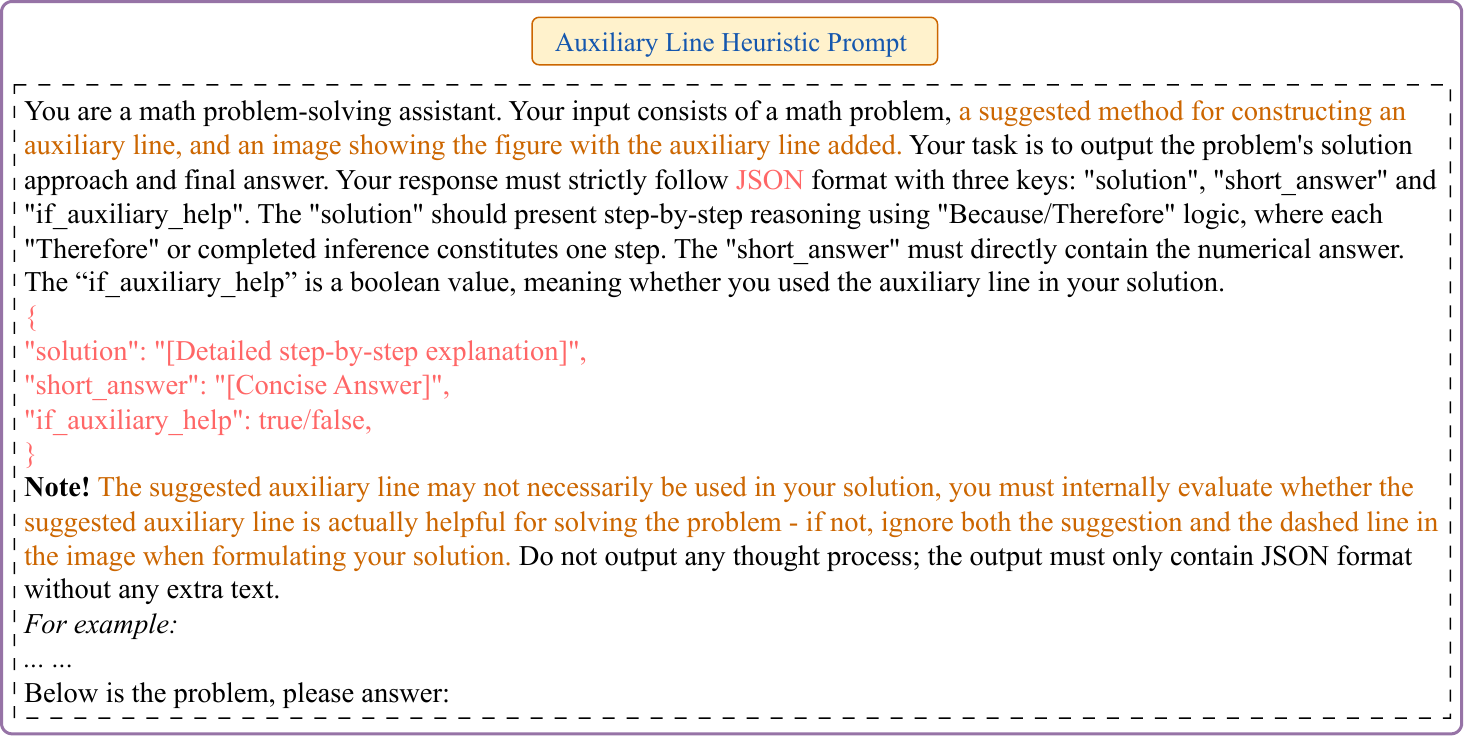} 
\caption{One-shot solution generation prompt for auxiliary line heuristic evaluaion.}
\label{fig:prompt_auxiliary}
\end{figure*}

In the auxiliary line heuristic experiment, we provide the LLM with both the auxiliary line construction method from the reference solution and the corresponding diagram showing this auxiliary line. The model is then prompted to analyze why this particular auxiliary line was suggested and determine whether to incorporate it into its own solution approach. The specific prompting methodology is illustrated in the accompanying Figure \ref{fig:prompt_auxiliary}.

\subsection{Prompt for Solution Evaluation.}
\label{sec:Evaluation_prompt}
\begin{figure*}[t!]   
\centering
\includegraphics[width=0.9\textwidth]{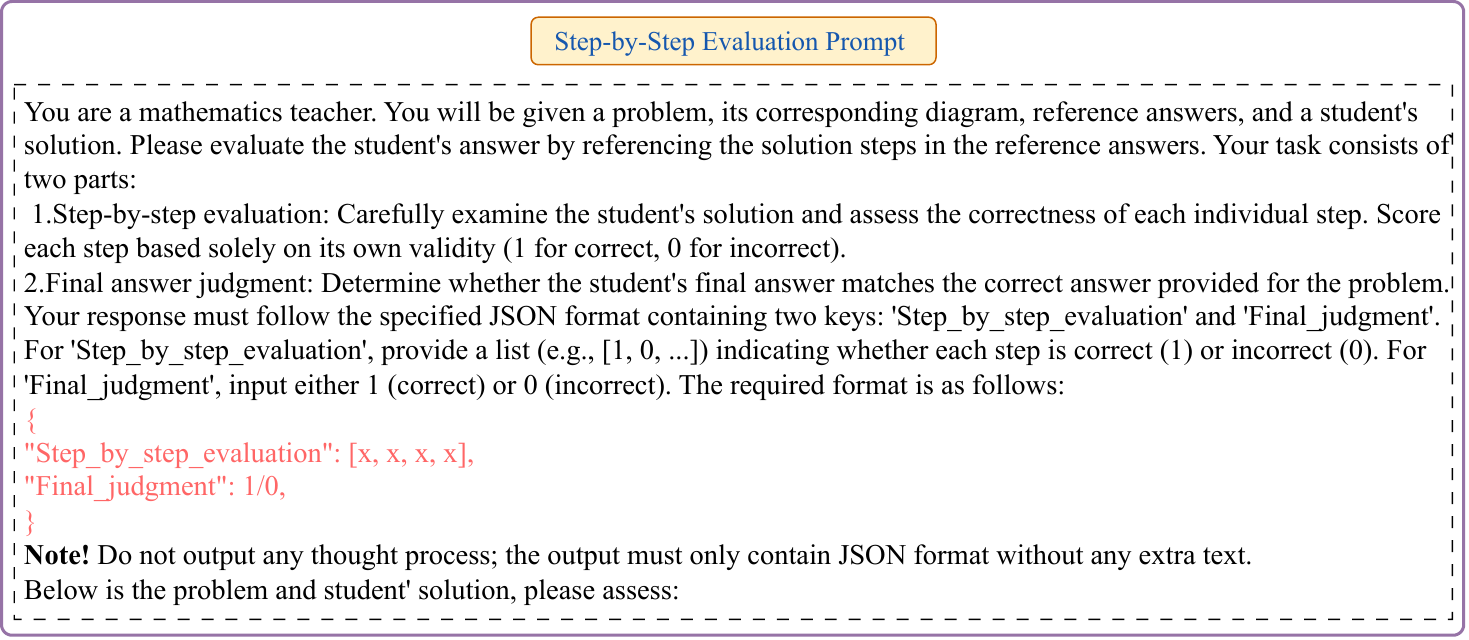} 
\caption{Zero-shot Step-by-Step Evaluation prompt.}
\label{fig:prompt_evaluation}
\end{figure*}

\begin{figure*}[t!]   
\centering
\includegraphics[width=0.9\textwidth]{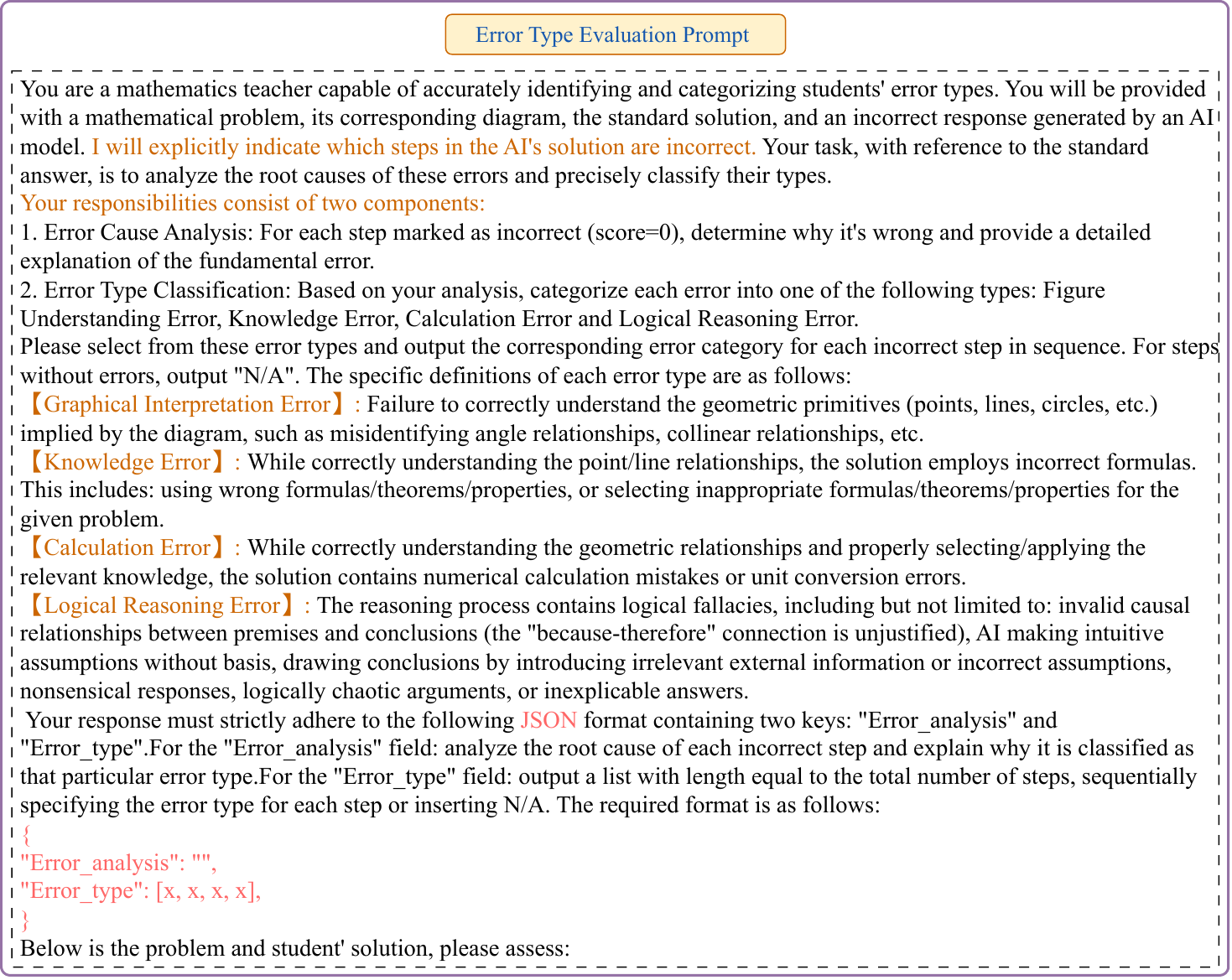} 
\caption{Zero-shot Error Type Evaluation prompt.}
\label{fig:prompt_errortype}
\end{figure*}

In our evaluation framework, we employ two distinct prompts to guide evaluators in assessing the generated solutions: one for step-by-step scoring and another for error type analysis, as illustrated in Figures \ref{fig:prompt_evaluation} and \ref{fig:prompt_errortype} respectively. Both assessment components are conducted with reference to the standard solution provided in the reference answers, thereby enhancing the reliability of our evaluation.

\subsection{Model Details.}

For the nine closed-source models, we access them through API and perform inference using simple CPU computation. For the four open-source models, we conduct inference using a server equipped with two NVIDIA A100 GPUs. The detailed generation parameters are specified in Table \ref{tab:model_hyperparams}.

\begin{table*}[h]
\centering
\setlength{\tabcolsep}{1mm}
\renewcommand{\arraystretch}{1.2} 
\scalebox{0.82}{
\begin{tabular}{l|l}
\hline
\textbf{Model} & \textbf{Hyperparameters} \\
\hline
GPT-4o & model = \texttt{gpt-4o-2024-08-06}, temperature = 0.1, max\_tokens = 4096 \\
GPT-4.1 & model = \texttt{gpt-4.1-2025-04-14}, temperature = 0.1, max\_tokens = 4096 \\
Claude-3.7 & model = \texttt{claude-3-7-sonnet-20250219}, temperature = 0.1, max\_tokens = 4096 \\
Claude-4.0 & model = \texttt{claude-sonnet-4-20250514}, temperature = 0.1, max\_tokens = 4096 \\
Claude-4.5 & model = \texttt{claude-sonnet-4-5-20250929}, temperature = 0.1, max\_tokens = 4096 \\
Claude-4.5-Thinking & model = \texttt{claude-sonnet-4-5-20250929-thinking}, temperature = 0.1, max\_tokens = 8192 \\
Gemini-2.0-Thinking & model = \texttt{gemini-2.0-flash-thinking-exp-01-21}, temperature = 0.1, max\_tokens = 8192 \\
Gemini-2.5-Pro & model = \texttt{gemini-2.5-pro-preview-03-25}, temperature = 0.1, max\_tokens = 10288 \\
o1 & model = \texttt{o1}, temperature = 0.1, max\_tokens = 8192 \\
o3 & model = \texttt{o3}, temperature = 0.1, max\_tokens = 10288 \\
o3-mini & model = \texttt{o3-mini-all}, temperature = 0.1, max\_tokens = 8192 \\
o4-mini & model = \texttt{o4-mini-2025-04-16}, temperature = 0.1, max\_tokens = 8192 \\
\hline
Qwen2.5-VL-72B & model = \texttt{Qwen/Qwen2.5-VL-72B-Instruct}, temperature = 0.1, max\_tokens = 10288 \\
Qwen2.5-VL-7B & model = \texttt{Qwen/Qwen2.5-VL-7B-Instruct}, temperature = 0.1, max\_tokens = 10288 \\

QvQ-72B & model = \texttt{Qwen/QVQ-72B-Preview}, temperature = 0.1, max\_tokens = 10288 \\
Qwen3-VL-32B & model = \texttt{Qwen/Qwen3-VL-32B-Instruct}, temperature = 0.1, max\_tokens = 10288 \\
Qwen3-VL-32B-Thinking & model = \texttt{Qwen/Qwen3-VL-32B-Thinking}, temperature = 0.1, max\_tokens = 10288 \\
GLM-4.1V-9B-Base & model = \texttt{ZhipuAI/GLM-4.1V-9B-Base}, temperature = 0.1, max\_tokens = 10288 \\
GLM-4.1V-9B-Thinking
 & model = \texttt{ZhipuAI/GLM-4.1V-9B-Thinking}, temperature = 0.1, max\_tokens = 10288 \\

InternVL3-78B & model = \texttt{OpenGVLab/InternVL3-78B}, temperature = 0.1, max\_tokens = 4096 \\
InternVL3-8B & model = \texttt{OpenGVLab/InternVL3-8B}, temperature = 0.1, max\_tokens = 4096 \\
InternVL3.5-38B & model = \texttt{OpenGVLab/InternVL3\_5-38B}, temperature = 0.1, max\_tokens = 4096 \\
InternVL3.5-8B & model = \texttt{OpenGVLab/InternVL3\_5-8B}, temperature = 0.1, max\_tokens = 4096 \\
\hline

\end{tabular}
}
\caption{Model Hyperparameters}
\label{tab:model_hyperparams}
\end{table*}

\section{Process Evaluation Cases} 
\label{appendix_evacase}


\subsection{Comparison of Different Evaluators} 
\begin{figure*}[t!]   
\centering
\includegraphics[width=1\textwidth]{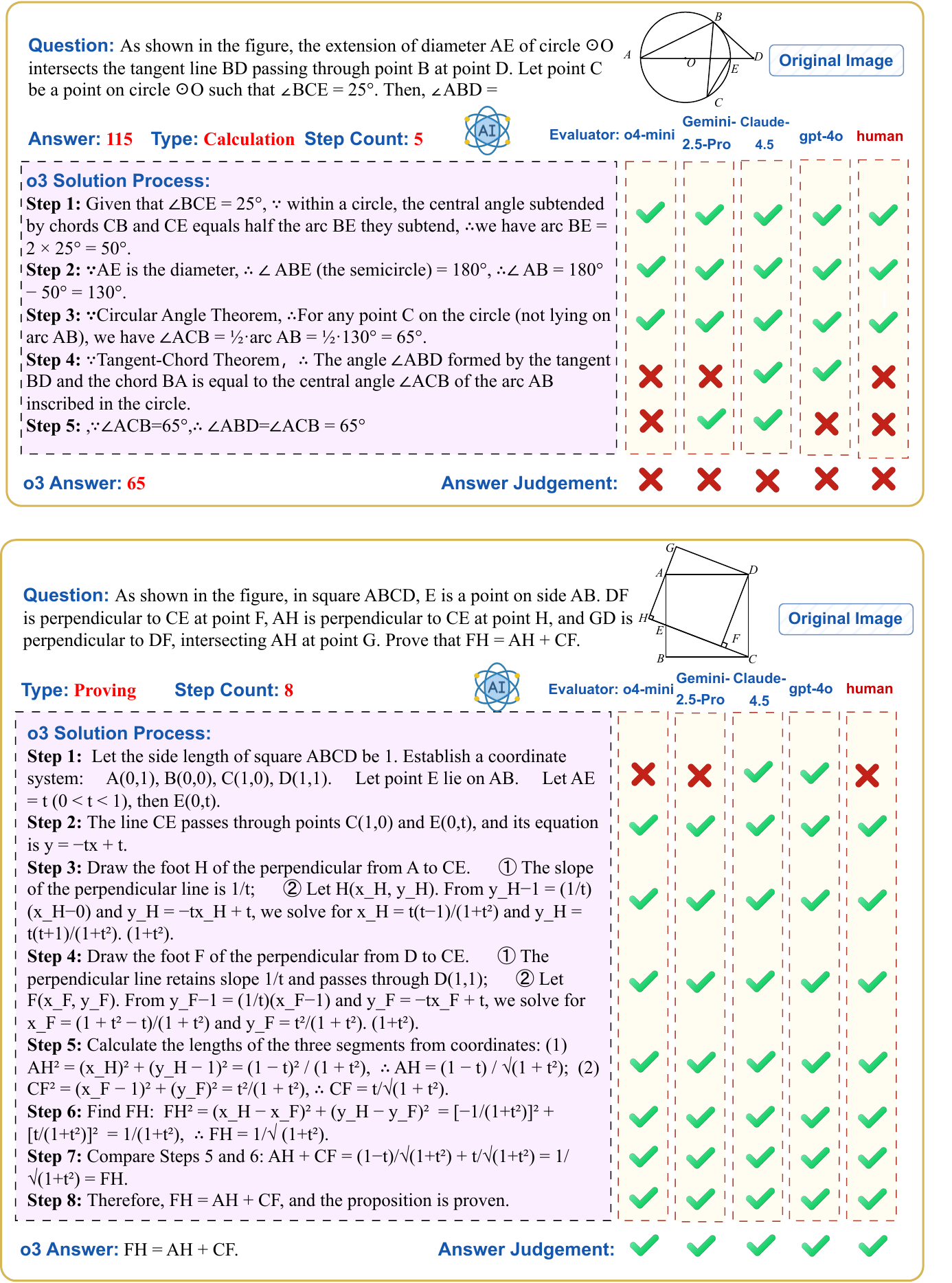} 
\caption{Comparative case of Evaluation Processes across Different MLLMs}
\label{fig:process_score}
\end{figure*}

Figure \ref{fig:process_score} presents the process scores assigned by four models on the same problem, alongside scores given by human evaluators. It can be seen that o4-mini's scoring for model solutions aligns most closely with human judgment, followed by Gemini 2.5 Pro. Notably, in some cases—such as the second example in the figure—for problems solved via coordinate system methods, human evaluators require complex computational verification, whereas the MLLM-based evaluator holds a distinct advantage in this regard.


\subsection{Different solution path evaluation cases}
\begin{figure*}[t!]   
\centering
\includegraphics[width=1\textwidth]{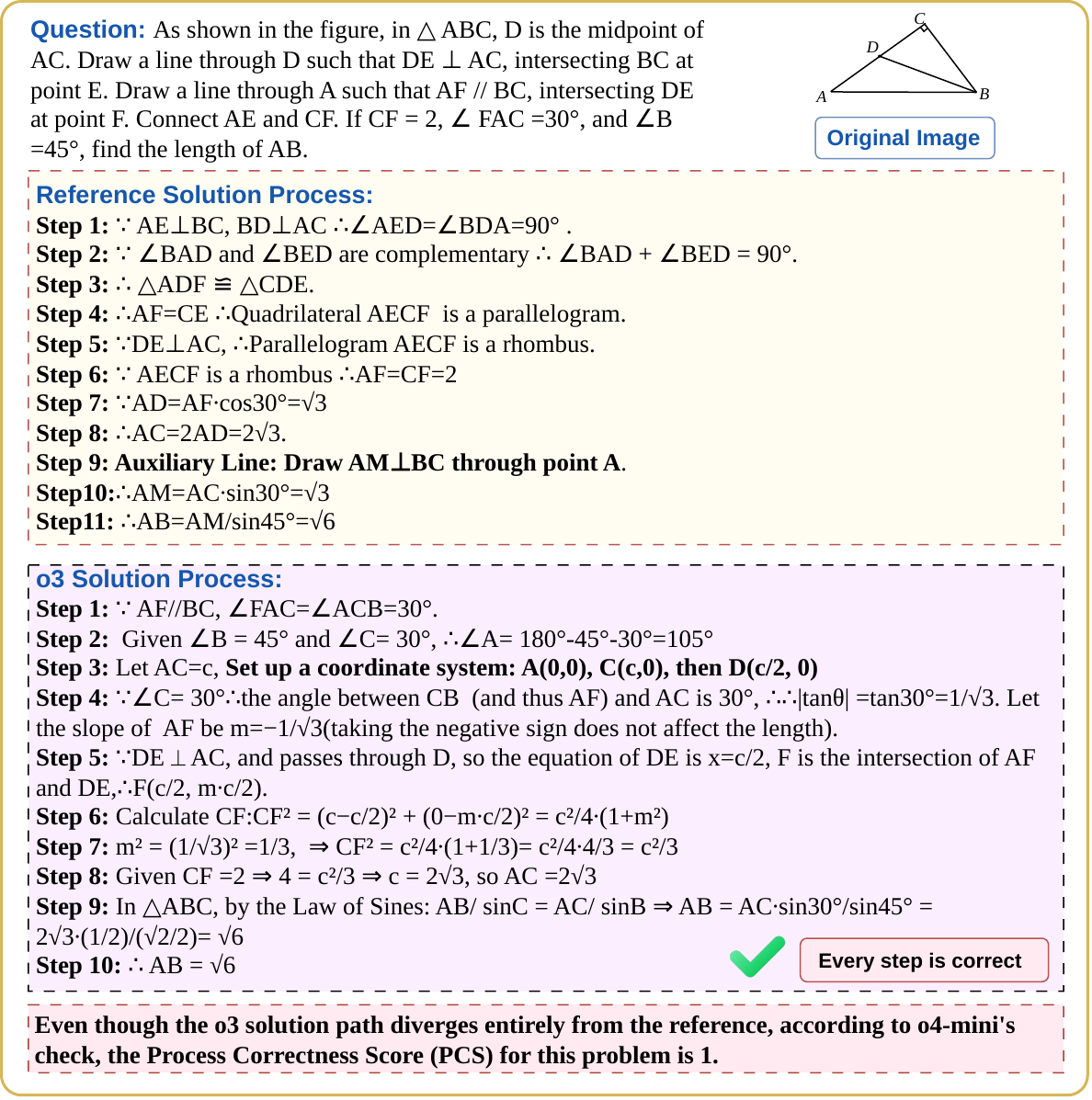} 
\caption{Evaluation case of Solutions Different from the Reference Answer}
\label{fig:different_solution}
\end{figure*}

Figure \ref{fig:different_solution} presents an evaluation case of an alternative solution path. In this example, while the reference solution employs auxiliary lines, the o3 model utilizes a coordinate geometry approach. Even though the o3 solution path diverges entirely from the reference, the evaluator (o4-mini) independently verifies the mathematical correctness and logic of each step. Since no conceptual, computational, or logical errors are found within the coordinate geometry framework, the evaluator identifies all steps as correct. Consequently, the Process Correctness Score (PCS) for this problem is 1.

\begin{figure*}[t!]   
\centering
\includegraphics[width=1\textwidth]{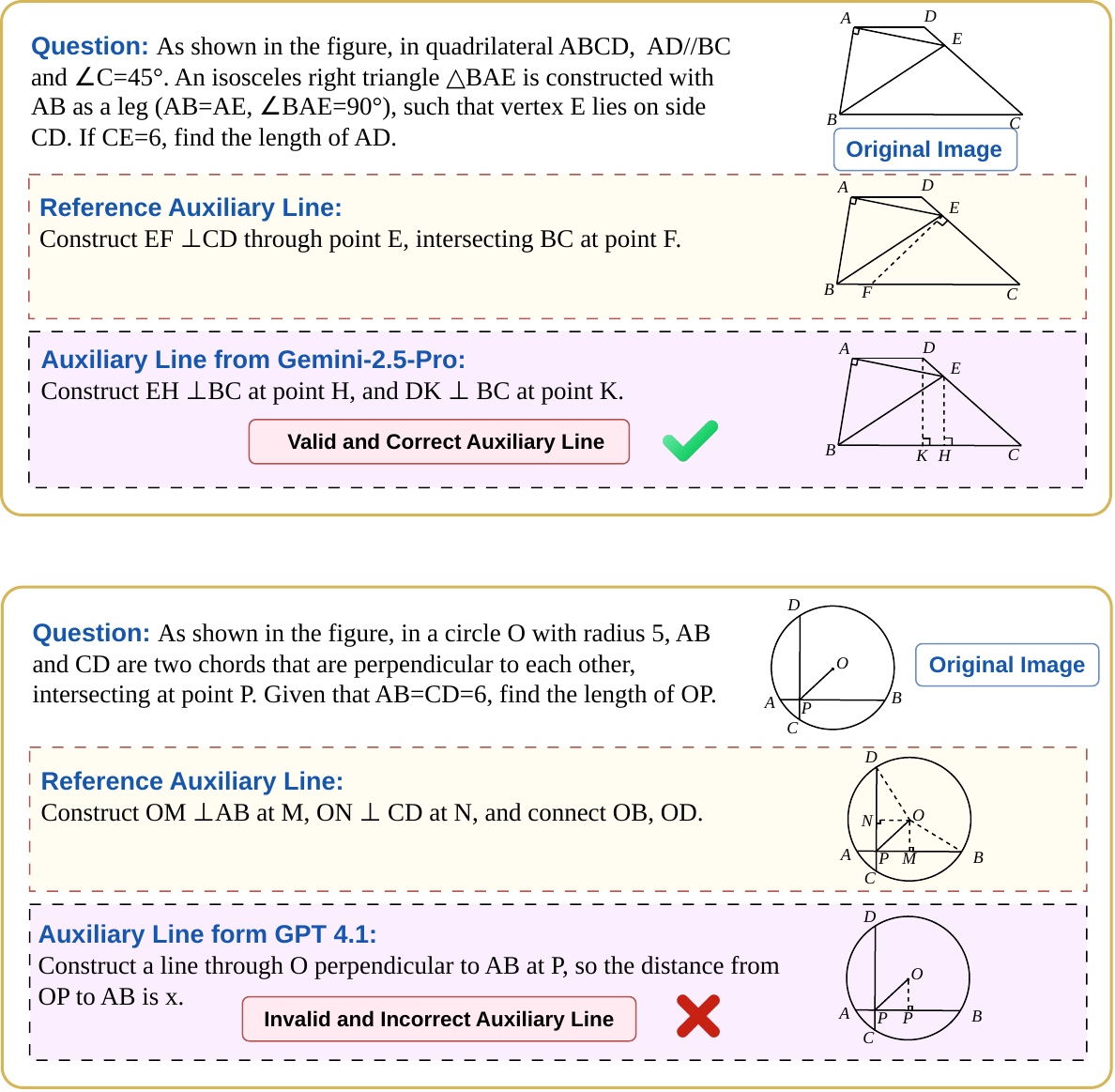} 
\caption{Evaluation case of Alternative Auxiliary Line Construction Methods}
\label{fig:different_auxiliary}
\end{figure*}

Regarding the auxiliary lines construction step, the evaluator judges whether this step conforms to geometric principles—distinguishing between valid, compliant constructions and impractical, invalid ones (e.g., "Construct $AD \perp AB$ with foot $D$" would be considered invalid because the notation "$AD \perp AB$" implies that the perpendicular intersection is at point "$A$", which contradicts the definition of "$D$" as the foot of the perpendicular). The evaluator marks auxiliary line construction that is geometrically sound as "correct" and subsequently assesses the validity of the steps that follow. Figure \ref{fig:different_auxiliary} presents two evaluation cases of alternative auxiliary lines: one involving a valid geometric relationship and another featuring an erroneous construction that violates geometric principles. The evaluator correctly assigns scores of 1 and 0, respectively.
\subsection{Proof Process Evaluation Cases} 
\begin{figure*}[t!]   
\centering
\includegraphics[width=0.98\textwidth]{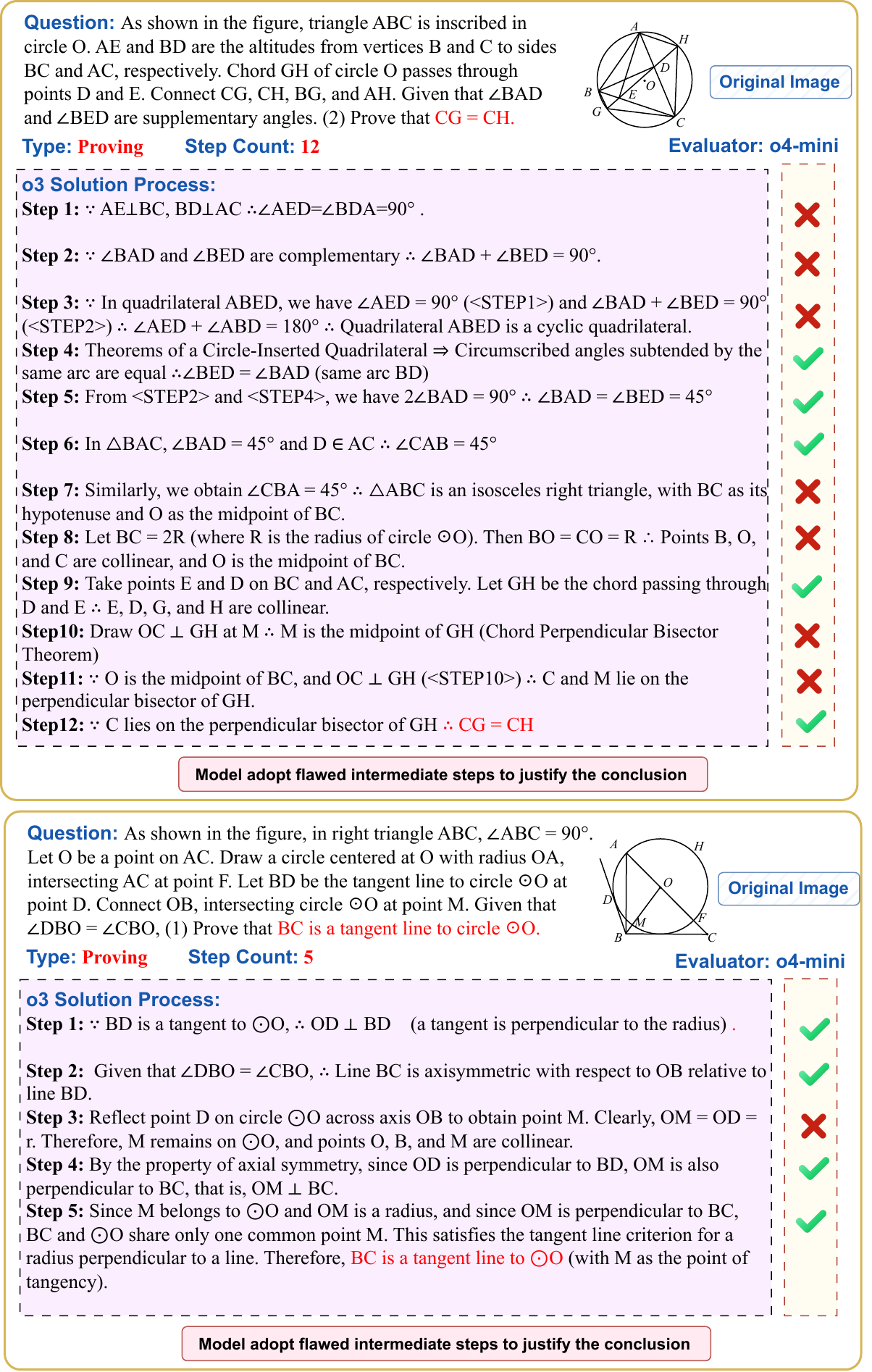} 
\caption{Examples of Shortcuts in Proof Problems}
\label{fig:proof_case}
\end{figure*}

Figure \ref{fig:proof_case} shows two proof problems in which models use intermediate steps with obvious errors to fit the final conclusion to be proved. It can be observed that the solution process is often absurd, committing egregious mistakes, revealing how models tend to cut corners when solving proof-based problems.

\begin{table*}[t]
    \centering
    \setlength{\tabcolsep}{1.5mm}
    \scalebox{0.85}{\begin{tabular}{@{} l | c c c c c@{}}
        \toprule
        \textbf{Model} & \textbf{1-4 Steps ACS} & \textbf{1-4 Steps PCS} & \textbf{5-8 Steps ACS} & \textbf{5-8 Steps PCS} & \textbf{Orig.Ans.Rate}\\
        \midrule
        Claude-4.5-T* & 89.7 / 85.4 (-4.3) & 68.6 / 70.1 (+1.5) & 82.3 / 79.8 (-2.5) & 55.8 / 54.7 (-1.1) & 1.3\% \\
        o3 & 94.5 / 95.2 (+0.7) & 84.1 / 85.5 (+1.4) & 94.7 / 94.2 (-0.5) & 80.2 / 75.8 (-4.4) & 1.3\%\\
        Gemini-2.5-Pro & 97.6 / 97.6 (+0.0) & 88.1 / 89.9 (+1.8) & 95.4 / 94.4 (-1.0) & 82.1 / 82.1 (+0.0) & \phantom{0.}0\%\\
    \midrule
        GPT-4.1 & 66.7 / 64.3 (-2.4) & 23.8 / 25.2 (+1.4) & 31.6 / 36.8 (+5.2) & 12.4 / 13.2 (+0.8) & 2.5\%\\
        Qwen3-VL-32B & 95.2 / 95.2 (+0.0) & 75.7 / 73.8 (-1.9) & 84.5 / 86.8(+2.3) & 50.5 / 51.2 (+0.7) & \phantom{0.}0\%\\
    \bottomrule
    \end{tabular}}
    \caption{Comparison of model performance before (left of /) and after (right of /) numerical perturbation. Values in "()" represent the change in performance when introducing numerical perturbations. "Orig. Ans. Rate (\%)" denotes the probability of the model incorrectly outputting the exact answer from the original, unperturbed problem.}
    \label{tab:numerical_perturbation}
\end{table*}
\section{Error Type Cases}
\label{appendix:error_type}
\begin{table}[h]
\centering
\scalebox{0.9}{
\begin{tabular}{lc}
\hline
Model & Avg. ROUGE-L ($\times10^{-2}$) \\
\hline
Gemini-2.5-Pro & 0.01 \\
Qwen3-VL-32B & 0.02 \\
Gemini-2.0-Thinking & 0.08 \\
o3 & 0.12 \\
o4-mini & 0.12 \\
o1 & 0.13 \\
\hline
\end{tabular}
}
\caption{ROUGE-L comparison across different models.}
\label{tab:rouge-l-scores}
\end{table}
The error types of the problems we used include four categories: (a) Figure Understanding Error, (b) Knowledge Error, (c) Calculation Error, and (d) Logical Reasoning Error. Their meanings are as follows:

\begin{enumerate}
    \item \textbf{Figure comprehension error:} Failure to correctly understand the geometric primitives (points, lines, circles, etc.) implied by the diagram, such as misidentifying angle relationships, collinear relationships, etc.
    \item \textbf{Knowledge Error:} While correctly understanding the point/line relationships, the solution employs incorrect formulas. This includes: using wrong formulas/theorems/properties, or selecting inappropriate formulas/theorems/properties for the given problem.
    \item \textbf{Calculation Error:} While correctly understanding the geometric relationships and properly selecting/applying the relevant knowledge, the solution contains numerical calculation mistakes or unit conversion errors.
    \item \textbf{Logical Reasoning Error:} The reasoning process contains logical fallacies, including but not limited to: invalid causal relationships between premises and conclusions (the "because-therefore" connection is unjustified), AI making intuitive assumptions without basis, drawing conclusions by introducing irrelevant external information or incorrect assumptions, nonsensical responses, logically chaotic arguments, or inexplicable answers.
\end{enumerate}

\begin{figure*}[t!]   
\centering
\includegraphics[width=0.98\textwidth]{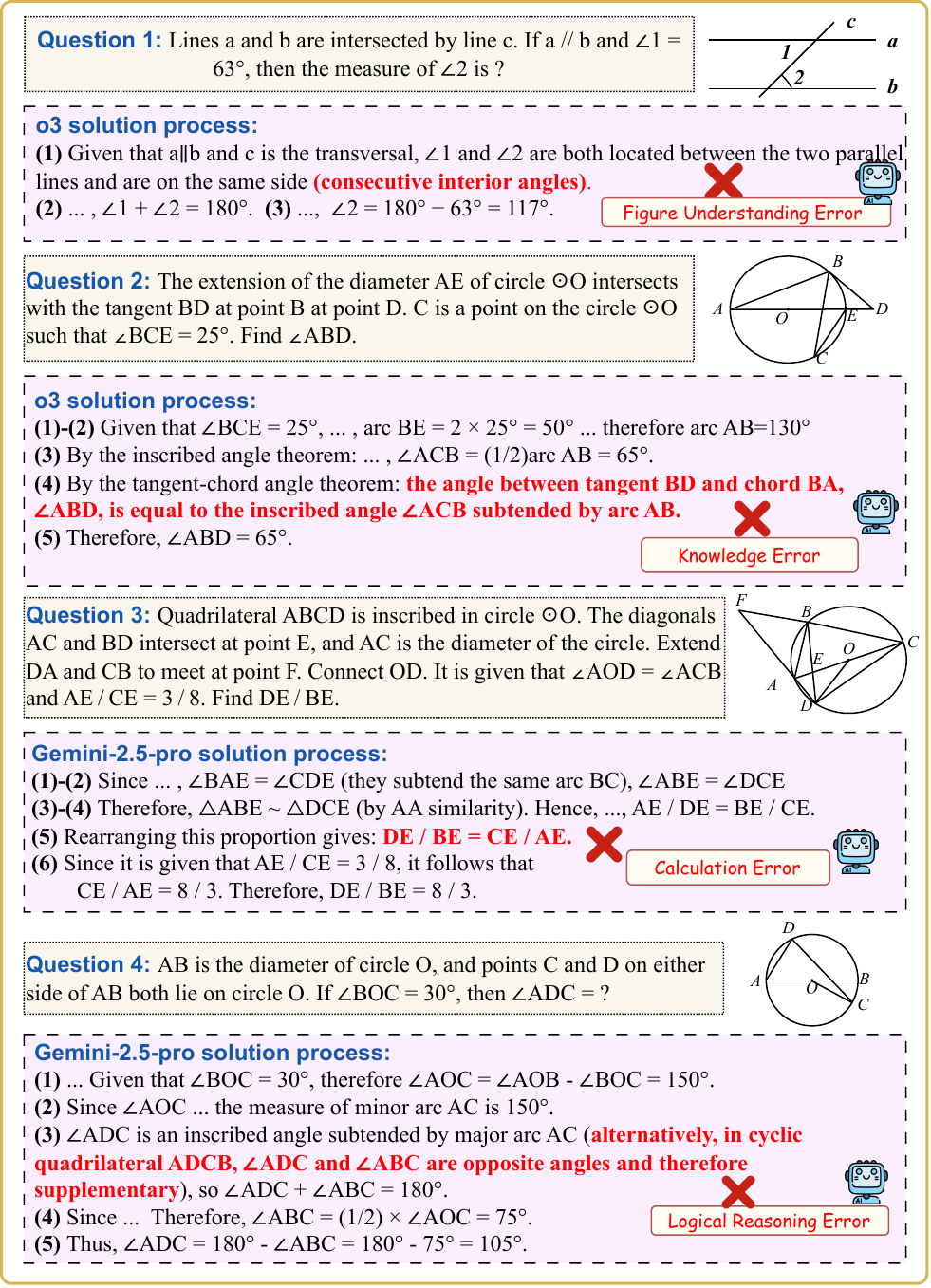} 
\caption{Examples of different error types.}
\label{fig:error_type_example}
\end{figure*}

A MLLM evaluator is employed with prompt in Appendix \ref{sec:Evaluation_prompt} to determine these error types and their corresponding steps. Some Error Cases are illustrated in Figure \ref{fig:error_type_example}.

\section{Data Leakage Analysis}
\label{appendix_Leakage}

\subsection{Textual Overlap Assessment}
\begin{figure*}[t!] 
\centering

\begin{minipage}[t]{0.48\textwidth}
    \centering
    \includegraphics[width=\textwidth]{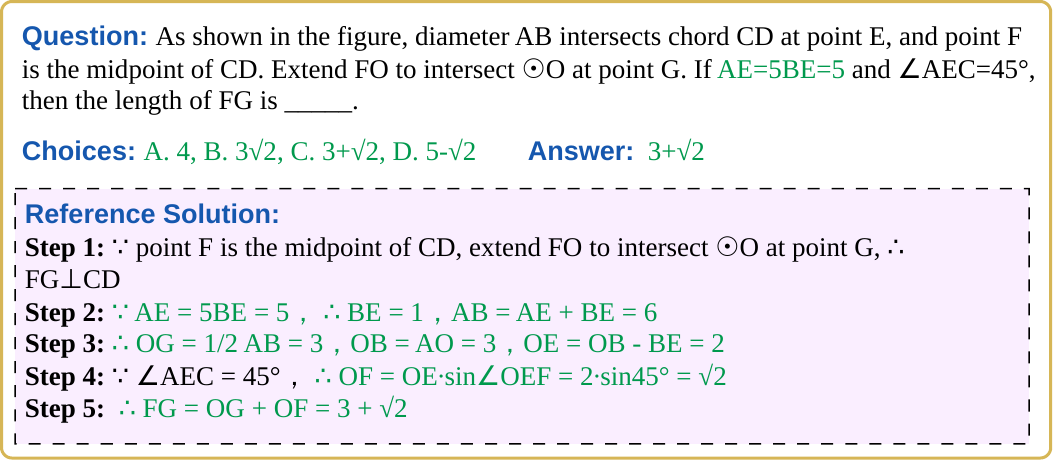}
    \vspace{0.5em} 
    {\small \textbf{(a)} Problem before perturbation}
    \label{fig:appendix_c_1}
\end{minipage}
\hfill
\begin{minipage}[t]{0.48\textwidth}
    \centering
    \includegraphics[width=\textwidth]{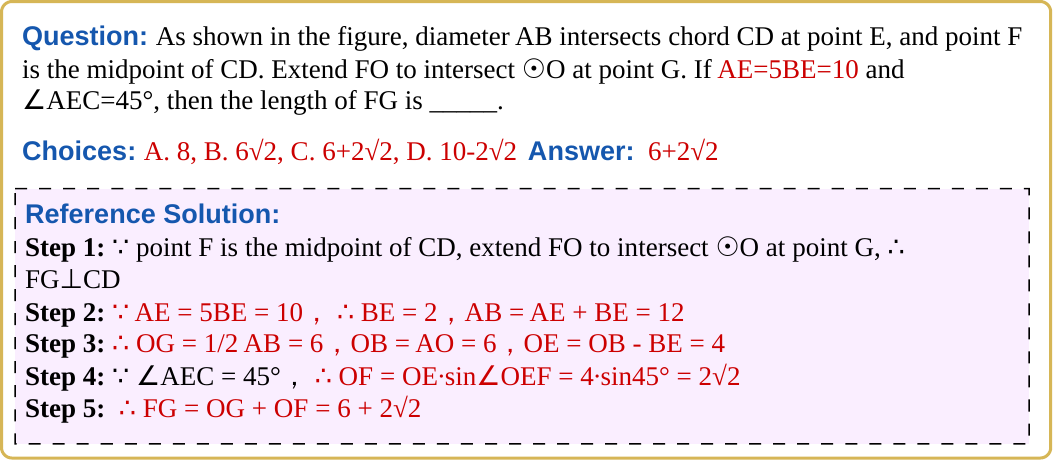}
    \vspace{0.5em}
    {\small \textbf{(b)} Problem after perturbation}
    \label{fig:appendix_c_proof}
\end{minipage}
\caption{Example of numerical perturbation.}
\label{fig:numerical_perturbation}
\end{figure*}

To evaluate potential data leakage in models while accounting for Chinese language characteristics, we conducted a similarity analysis using a two-step approach. We randomly selected 150 problems from our dataset and generated 3 distinct prompts for each problem to query the models, resulting in a total of 450 test cases. First, we generated responses using a 10-gram matching process, where each Chinese character was treated as a token. Subsequently, we employed the Chinese-adapted ROUGE-L metric to measure the overlap between these model outputs and ground-truth solutions.
We observed uniformly low similarity metrics across all models (Table \ref{tab:rouge-l-scores}), with average ROUGE-L scores ranging from $0.01 \times 10^{-2}$ to $0.13 \times 10^{-2}$ (0.01\% to 0.13\%). In particular, these same models exhibited poor matching accuracy in our contamination detection experiments, despite their strong general reasoning capabilities. Similar to the approach in previous works \citep{xu2024benchmarking,gao2024omni}, our results suggest that these models have not been exposed to or trained on our dataset. The consistently low ROUGE-L scores, all averaging below $0.14 \times 10^{-2}$ (0.14\%), indicate minimal overlap between our dataset and the training corpora of the tested models. These scores fall well within the expected range for clean, unleaked data, further confirming the novelty and integrity of our dataset.

\subsection{Numerical Perturbation Test}
Beyond n-gram analysis, to more reliably and definitively verify whether models are genuinely reasoning rather than memorizing, we designed a numerical perturbation experiment.

We focused on 159 short- and medium-step calculation problems in the GeoLaux-mini set, as models generally perform better on these shorter problems, any reliance on memorization (contamination) would be much more obvious. Specifically, we utilized the advanced Gemini-3-Pro to apply random numerical perturbations to these 159 problems, followed by strict human verification. We ensured that only the specific numerical values are altered, while all geometric relationships and the core questions remained completely identical.

As shown in Table \ref{tab:numerical_perturbation}, the models' performance shows no significant change after introducing numerical perturbations, and they almost never output the original, pre-perturbation answers (any rare occurrences are likely due to general solving errors). This provides strong evidence that the models are engaging in genuine reasoning rather than simply outputting memorized data from pre-training. Furthermore, it underscores the integrity and practical significance of our dataset, confirming it is free from data leakage and ensures a fair evaluation. Figure \ref{fig:numerical_perturbation} is an example before and after the disturbance. This perturbed dataset has also been made publicly available on GitHub.


\end{document}